\documentclass[times,referee,twocolumn,final,authoryear]{elsarticle}

\usepackage{ycviu}
\usepackage{framed,multirow}

\usepackage{amssymb}
\usepackage{latexsym}

\usepackage{amsmath}    
\usepackage{adjustbox}  
\usepackage{booktabs}   

\usepackage{url}

\usepackage[table]{xcolor}
\definecolor{rblue}{rgb}{0,0.5,1}

\journal{Computer Vision and Image Understanding}

\begin{document}

\begin{frontmatter}

\title{Exploring Event-based Human Pose Estimation with 3D Event Representations}
  
\author[1]{Xiaoting Yin*}
\author[1,4]{Hao Shi*}
\cortext[]{*Equal contribution}
\author[1]{Jiaan Chen*}
\author[1]{Ze Wang}
\author[1]{Yaozu Ye}
\author[2,3]{Kailun Yang\corref{cor2}}
\cortext[cor2]{Corresponding author 2: Tel.: +86-0731-88821990}
\ead{kailun.yang@hnu.edu.cn}
\author[1]{Kaiwei Wang\corref{cor1}}
\cortext[cor1]{Corresponding author 1: Tel.: +86-0571-87953154}
\ead{wangkaiwei@zju.edu.cn}

\address[1]{State Key Laboratory of Extreme Photonics and Instrumentation, Zhejiang University, Hangzhou 310027, China}
\address[2]{School of Robotics, Hunan University, Changsha 410012, China}
\address[3]{National Engineering Research Center of Robot Visual Perception and Control Technology, Hunan University, Changsha 410082, China}
\address[4]{Shanghai SUPREMIND Technology Company Ltd., Shanghai 201210, China}

\begin{abstract}
Human pose estimation is a fundamental and appealing task in computer vision. Although traditional cameras are commonly applied, their reliability decreases in scenarios under high dynamic range or heavy motion blur, where event cameras offer a robust solution. Predominant event-based methods accumulate events into frames, ignoring the asynchronous and high temporal resolution that is crucial for distinguishing distinct actions. To address this issue and to unlock the 3D potential of event information, we introduce two 3D event representations: the Rasterized Event Point Cloud (RasEPC) and the Decoupled Event Voxel (DEV). The RasEPC aggregates events within concise temporal slices at identical positions, preserving their 3D attributes along with statistical information, thereby significantly reducing memory and computational demands. Meanwhile, the DEV representation discretizes events into voxels and projects them across three orthogonal planes, utilizing decoupled event attention to retrieve 3D cues from the 2D planes. Furthermore, we develop and release EV-3DPW, a synthetic event-based dataset crafted to facilitate training and quantitative analysis in outdoor scenes. Our methods are tested on the DHP19 public dataset, MMHPSD dataset, and our EV-3DPW dataset, with further qualitative validation via a derived driving scene dataset EV-JAAD and an outdoor collection vehicle. Our code and dataset have been made publicly available at \url{https://github.com/MasterHow/EventPointPose}.
\end{abstract}

\end{frontmatter}

\begin{figure}[!t]
\centering
\includegraphics[width=1.0\linewidth]{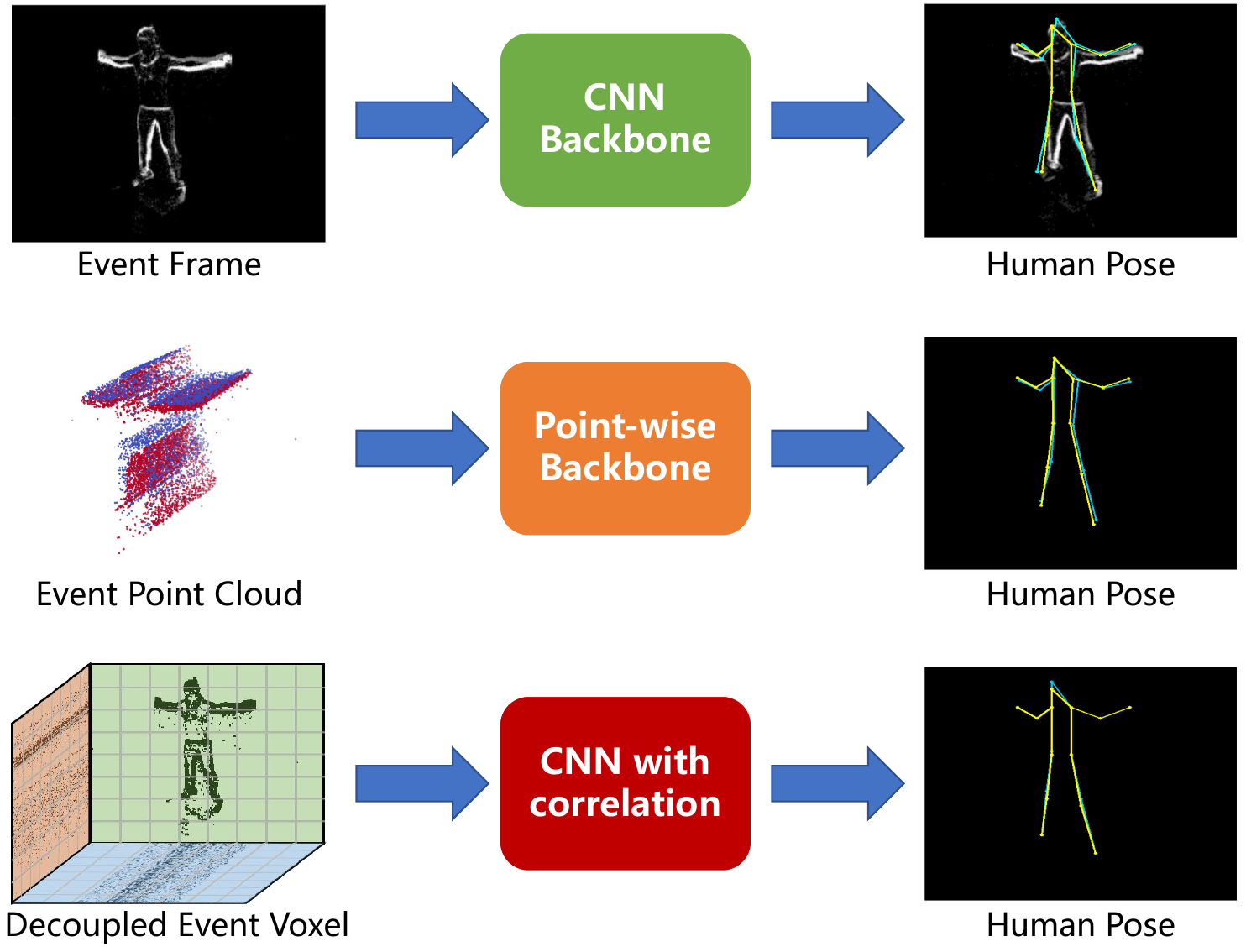}
\caption{2D event frame based human pose estimation paradigm \textit{vs.} the proposed 3D Rasterized Event Point Cloud (RasEPC) based paradigm \textit{vs.} the proposed novel 3D Decoupled Event Voxel (DEV) based paradigm.
}
\label{intro_2}
\end{figure}

\section{Introduction}
Human Pose Estimation (HPE) is a vital task in computer vision, aiming to accurately localize human anatomical keypoints with a wide range of applications such as action recognition~\citep{yan2018stgcn, shi2019skelaction, yang2020hierarchical}, animations~\citep{willett2020pose2pose}, and autonomous driving~\citep{cadena2019pedestrian}.
Traditional HPE approaches primarily rely on standard RGB cameras, but they often struggle in complex scenarios with fast motion and large dynamic range. 
Yet, event cameras with high temporal resolution and large dynamic range can maintain stable signal output under the above challenging circumstances~\citep{gallego2020eventsurvey}.

Recently, many studies~\citep{calabrese2019dhp19, scarpellini2021lifting} based on event cameras process event information as frame-like images, neglecting the valuable temporal order of events.
Yet, this unique high-temporal-resolution feature of event cameras is particularly beneficial for interpreting complex human body movements.
For example, when raising a hand, the event stream captures not only the movement but also the precise temporal sequence of these actions, allowing for real-time analysis of the arm's upward trajectory.
However, when the event stream is aggregated into frames, this inherent temporal information is lost.
While event frame is widely used for embedding into neural networks, it eliminates the fine-grained timing details essential for distinguishing distinct actions.
As a result, actions like raising and lowering the hand, which are intrinsically characterized by their temporal order, may appear remarkably similar.
This similarity poses a challenge for temporally aggregated event representations to effectively differentiate between these actions.

To tackle the above problem, we aim to explore the event representation design and contribute an alternative perspective centered on the human pose estimation task. 
Events are typically represented as a set of \(N\) input events \(\{(x_i, y_i, t_i, p_i)\}_{i \in [1, N]}\), which can be viewed as scattered points in the three-dimensional space \([x, y, t]\).
Similarly, in LiDAR-related fields, three-dimensional data is often represented as a set of points in the \([x, y, z]\) space, each characterized by its coordinates and \(n\) additional features.
This analogy inspires us to leverage mature techniques from LiDAR to process event-based data. In 3D data processing, point clouds and voxel grids are common representation methods. When applied to event data, these methods retain characteristics of asynchronous sparsity and high temporal resolution.

In this work, we propose two complementary 3D event representations: the Rasterized Event Point Cloud (RasEPC), inspired by point clouds, and the Decoupled Event Voxel (DEV), inspired by voxel grids.
Both of the proposed 3D event representations are primarily designed for single-person 2D human pose estimation.
They also offer flexible adaptability to 3D human pose estimation, exemplified by the DHP19 dataset, through the incorporation of multi-view camera systems. 
This flexibility is further extended to encompass multi-person pose estimation scenarios by integration with advanced detection algorithms, as evidenced by our work on both the EV-3DPW and EV-JAAD datasets, as well as in applications involving outdoor collection vehicles.

Events are inherently asynchronous and sparse, and using sparse representations like point clouds can further enhance their efficiency.
To verify the generality of our proposed framework, we implement event point clouds on three prevalent point cloud networks.
While point cloud frameworks have been widely used in the field of LiDAR point clouds~\citep{qi2017pointnet, wang2019dgcnn, zhao2021pointtrans}, they have not been fully studied in event-based HPE. 
This is due to the additional challenges posed by the large number of events in the event stream.
Given the impressive microsecond-level time resolution of event cameras, using all recorded events can significantly slow down forward propagation and lead to high memory consumption. 
Therefore, we propose the Rasterized Event Point Cloud (RasEPC) representation, which reduces the number of events while preserving their 3D features.

Additionally, we propose the Decoupled Event Voxel (DEV) representation, placing events into a grid and projecting them onto three orthogonal planes. 
The voxel-based representation usually has large memory and computational consumption, which is contrary to the original intention of using event cameras for real-time HPE prediction, so we decouple the voxel into three orthogonal projections. 
Although implicit scene representation has been applied to 3D perception~\citep{huang2023tri}, its suitability for event data and the extraction of spatial-temporal features from 2D planes remain unexplored.
The incorporation of our Decoupled Event Attention (DEA) module, which injects spatial-temporal information into the implicit representation, facilitates the extraction of the 3D cues from the tri-view 2D planes, making it effective for event-based HPE.

Aside from considering the inherent properties of event-based data in our architectural design, we also address the shortage of available event training data by establishing and releasing a new synthetic event-based HPE benchmark, EV-3DPW. 
This challenging dataset features multiple people interacting with the environment, fostering in-the-wild research on HPE.
We generate the dataset through the ESIM simulator~\citep{rebecq2018esim_simulator} and DVS-Voltmeter~\citep{lin2022dvsvoltmeter}. 
3DPW~\citep{von2018recovering} consists of more than $51,000$ frames in challenging sequences, including walking in the city, going upstairs, having coffee, or taking the bus.
We further generate the EV-JAAD dataset, derived from the driving scene dataset JAAD~\citep{rasouli2017they}, for testing the generalization performance of different representation strategies in unseen scenes.

We conduct extensive quantitative experiments on the real-world collected event dataset DHP19~\citep{calabrese2019dhp19}, challenging public dataset MMHPSD~\citep{zou2021eventhpe} and our simulated EV-3DPW dataset. 
The RasEPC representation on three datasets accomplishes real-time prediction speeds on mobile devices with satisfactory accuracy, while the DEV representation obtains the highest accuracy under the same backbone.
On the established street scene dataset EV-3DPW, significant differences exist between the training and test sets, including variations in background complexity, types of actions, and clothing colors. 
Methods based on event data outperform those based on RGB images using the same backbone, primarily because they capture changes in brightness.
We further leverage models trained on the simulated EV-3DPW dataset for zero-shot testing on the EV-JAAD dataset and an outdoor collection vehicle to assess the practical applicability and generalization of simulated data to real-world unseen scenarios.
The two 3D event representations we proposed focus on placing event information in the 3D space of $[x, y, t]$, instead of the traditional method of using the 2D space of $[x, y]$ with $t$ in the feature dimension. 
Consequently, both 3D event representations better maintain the high temporal resolution characteristics of event cameras.
In the task of event-based human pose estimation, we verify that RasEPC, which has undergone rasterization preprocessing, can achieve the lowest latency with comparable accuracy on edge computing platforms in the form of sparse data. 
The DEV representation, incorporating the Decoupled Event Attention module, captures sufficient neighborhood information to achieve more accurate key point estimation than the two-dimensional event frame under the same backbone.

At a glance, we deliver the following contributions:
\begin{enumerate}
    \item We reformulate the event-based HPE problem from the perspective of three-dimensional event representation, offering an alternative to the dominant design of accumulating events to two-dimensional event frames.
    \item We exploit 3D event point clouds directly to demonstrate the feasibility of applying the well-known LiDAR point cloud learning backbones to human pose estimation.
    \item We extract and fuse decoupled event voxel features by integrating the conventional 2D CNN backbone with our Decoupled Event Attention Module to facilitate precise keypoint regression.
    \item We introduce a new, synthetic dataset EV-3DPW in the wild for investigating event-based HPE in street scenes.
\end{enumerate}

This paper is an extension of our conference work~\citep{chen2022efficient}. Within this publication, we significantly increase the insights into the task of event-based human pose estimation by adding the following contributions:
\begin{enumerate}
    \item We propose a novel 3D event representation that decouples event voxels to three orthogonal dimensions, reducing memory and computing consumption while retaining 3D information.
    \item We introduce a Decoupled Event Attention (DEA) module, offering a flexible strategy to effectively retrieve 3D cues from the tri-view DEV representation.
    \item We generate EV-3DPW, a new publicly available event-based HPE dataset that consists of diverse human activities in outdoor scenes, providing pairs of RGB images and events, along with human annotation boxes and human pose estimation labels.
    \item Our final DEV model achieves the best results under the same backbone on the public DHP19 dataset, MMHPSD dataset, and our simulated EV-3DPW dataset.
    \item Both proposed 3D event representations demonstrate strong generalization ability in real-world unseen scenarios on the synthetic EV-JAAD dataset and our captured outdoor event streams.
\end{enumerate}

\section{Related Work}
\subsection{Human Pose Estimation}
In recent years, Human Pose Estimation (HPE) has become a very important topic in the field of computer vision, mainly based on images and videos. 
Generally, it is categorized into 2D and 3D HPE.
In the field of 2D HPE, most research works~\citep{wei2016convposemachain, xiao2018simbase} deploy a Convolutional Neural Network (CNN) model and utilize the concept of heatmap to obtain the most likely keypoint coordinates, typically outperforming direct prediction methods~\citep{toshev2014deeppose}.
A multitude of studies further explore multi-person human pose estimation solutions, which generally include two strategies: top-down~\citep{fang2017rmpe, xiao2018simbase} and bottom-up~\citep{cao2017openpose, GengSXZW21DEKR}.
Top-down methods first detect the bounding boxes of individual persons and then estimate keypoints within these boxes, often yielding higher accuracy. 
Conversely, bottom-up methods commence by detecting all keypoints across the entire image and subsequently segregating multiple human body instances.

A host of approaches estimate 3D HPE via single cameras~\citep{Chen_2017_CVPR3d2dmatch, liu2022recent, sun2023trace} or multiple cameras~\citep{dong2019fast}. 
Monocular solutions, limited by a single camera's ability to capture depth, face challenges such as occlusions and ambiguities~\citep{scarpellini2021lifting}, while multi-camera setups can help overcome these issues.
Multi-view-based 3D HPE is often implemented based on matching and triangulation reconstruction~\citep{dong2019fast}.
In some mobile application scenarios, real-time performance becomes a key indicator, and knowledge distillation is often used to obtain small models~\citep{zhang2019fasthpe}. 
In this work, to facilitate a fair comparison, we do not use model compression techniques on any model. 
Our fundamental approach is centered around single-person 2D HPE, offering the flexibility to extend to 3D human pose estimation by incorporating multi-view cameras or to multi-person pose estimation by integrating with detection algorithms.

\subsection{Event-based Human Pose Estimation}
Event cameras are sensitive to moving objects~\citep{gallego2020eventsurvey}, so they have natural advantages in fields related to human motion~\citep{chen2022efficient, wu2020multipathhumanaction}.
Enrico~\textit{et al.}~\citep{calabrese2019dhp19} present the first event-based HPE dataset and introduce a lightweight CNN to estimate 2D key points, which are then triangulated into 3D poses.
EventCap~\citep{xu2020eventcap} takes both asynchronous event streams and low-frequency grayscale intensity images as input for 3D human motion capture using a single event camera.
LiftMono-HPE~\citep{scarpellini2021lifting} employs 2D event frames to estimate poses combined with implicit depth estimation to achieve 3D HPE through a single camera.
EventHPE~\citep{zou2021eventhpe} proposes a two-stage deep learning method that combines optical flow estimation and provides a home-grown dataset, MMHPSD. 
TORE~\citep{baldwin2022TORE} obtains very competitive 3D results on the DHP19 dataset by storing raw spike timing information and is not suitable for a fair comparison of our methods with a temporal process. 
In other event-based areas, HATS~\citep{sironi2018hats} represents events as histograms of average time surfaces but discards temporal information; Ev-FlowNet~\citep{zhu2018ev} uses a four-dimensional grid but loses earlier timestamps; UnFlow~\citep{zhu2019unsupervised} creates an event volume by discretizing time into $B$ bins, and when this volume is fed into the network, the time dimension is treated as the $C$ channel, limiting interaction between bins; and EST~\citep{gehrig2019end} learns representations end-to-end but is prone to overfitting and sensitive to event density. 
Unlike these approaches, we focus on preserving comprehensive temporal information while effectively capturing spatial and structural features, specifically for event-based human pose estimation.
VMST-Net~\citep{liu2023voxel} utilizes a transformer-based architecture for HPE with event voxels.
While both VMST-Net and our DEV method start by placing event streams into a voxel grid, DEV enhances the efficiency and further emphasizes 3D event neighbor correlation via implicitly representing event voxel through projections in three orthogonal directions and effective retrieval by decoupled event attention.

In the field of event vision, the DHP19 dataset~\citep{calabrese2019dhp19} is the first to capture human pose estimation using event cameras.
Recently, a large-scale multimodal event-based pose dataset, named CDEHP~\citep{shao2024temporal}, has been proposed.
When the number of event-based datasets is limited, plenty of studies generate datasets through event simulators~\citep{gehrig2020video_to_events,rebecq2018esim_simulator}.
EventGAN~\citep{zhu2021eventgan} demonstrates that event-based HPE datasets generated by the model can be used for training, seamlessly generalizing to real-world data. 
This approach has also been applied to event-based semantic segmentation~\citep{zhang2021exploring_event_segmentation}.
Scarpellini~\textit{et al.}~\citep{scarpellini2021lifting} produce a new dataset of simulated events from the standard RGB Human3.6m dataset. 
We note that, until now, there is a dearth of available datasets targeting outdoor complex street scenes with moving event cameras for training and evaluation.
This work attempts to fill this gap by creating synthetic datasets derived from the existing RGB outdoor 3DPW dataset~\citep{von2018recovering}.
Our EV-3DPW dataset serves as a valuable resource for training models in complex outdoor scenes, fostering research and development in the event vision area.

\subsection{Point Clouds \textit{vs.} Voxel Grid}
When dealing with LiDAR-related vision tasks, 3D data are often represented as point clouds~\citep{qi2017pointnet, wang2019dgcnn, zhao2021pointtrans} or voxel grids~\citep{wu20153d, maturana2015voxnet, qi2016volumetric}.
A point cloud is represented as an unordered set of 3D points, where each point is a vector of its 3D coordinate plus extra feature channels. 
Similarly, in event-based tasks, events can be interpreted as scattered points in the $[x,y,t]$ space, analogous to the $[x,y,z]$ representation in LiDAR data.
PointNet~\citep{qi2017pointnet} takes point clouds as direct input with respect to the permutation invariance of points. 
DGCNN~\citep{wang2019dgcnn} enhances its representation power by restoring the topological information of point clouds. 
Point transformer~\citep{zhao2021pointtrans} applies a self-attention-based network. 
As a sparse data form, point clouds are very useful in real-time applications, but the spatial neighboring relations between points are discarded. 
Another efficient way to perceive a 3D scene is to discretize the 3D space into voxels and assign a vector to represent each voxel. 
Voxelization describes fine-grained aggregated 3D structures including valuable neighboring information, but it is constrained by the computation cost of 3D convolution.

As another kind of three-dimensional data, the 3D representation of events has not been fully explored. 
Some related works~\citep{wang2019space, chen2020eventdgcnn} on human action recognition have explored the possibilities of using raw event points, and our preliminary conference work~\citep{chen2022efficient} has introduced the rasterized event point cloud representation in the unexplored area of human pose estimation, as shown in Fig.~\ref{intro_2}.
Our previous work demonstrates the efficiency of the event point cloud processing paradigm and shows its real-time advantages on edge computing platforms. 
Yet, due to the loss of geometric neighboring information, event point cloud methods encounter challenges in enhancing accuracy.
As aforementioned, the ability to describe the aggregated neighbor context of 3D structures makes voxel-based representation favorable for LiDAR-centric surrounding perception~\citep{huang2023tri}. 
While there are studies in the event vision field that have harnessed event voxels for tasks like optical flow and depth estimation~\citep{zhu2019unsupervised, hidalgo2020learning}, it remains scarcely considered for human pose estimation. 
The closest work to ours is VMST-Net~\citep{liu2023voxel}, which builds upon the prior work~\citep{chen2022efficient} and seeks to utilize 3D event voxels for HPE. 
Instead, our approach goes a step further by decomposing 3D event voxel data into three mutually orthogonal directions and inferring human joint locations through a fusion of established 2D CNN backbones and our novel decoupled event attention mechanism. 
This enables us to enhance the efficiency of voxel-based representations.
To the best of our knowledge, we are the first to use implicit representation to model event information for event-based human pose estimation.

\section{Methodology}
\begin{figure*}[!t]
\centering
\includegraphics[width=1.0\linewidth]{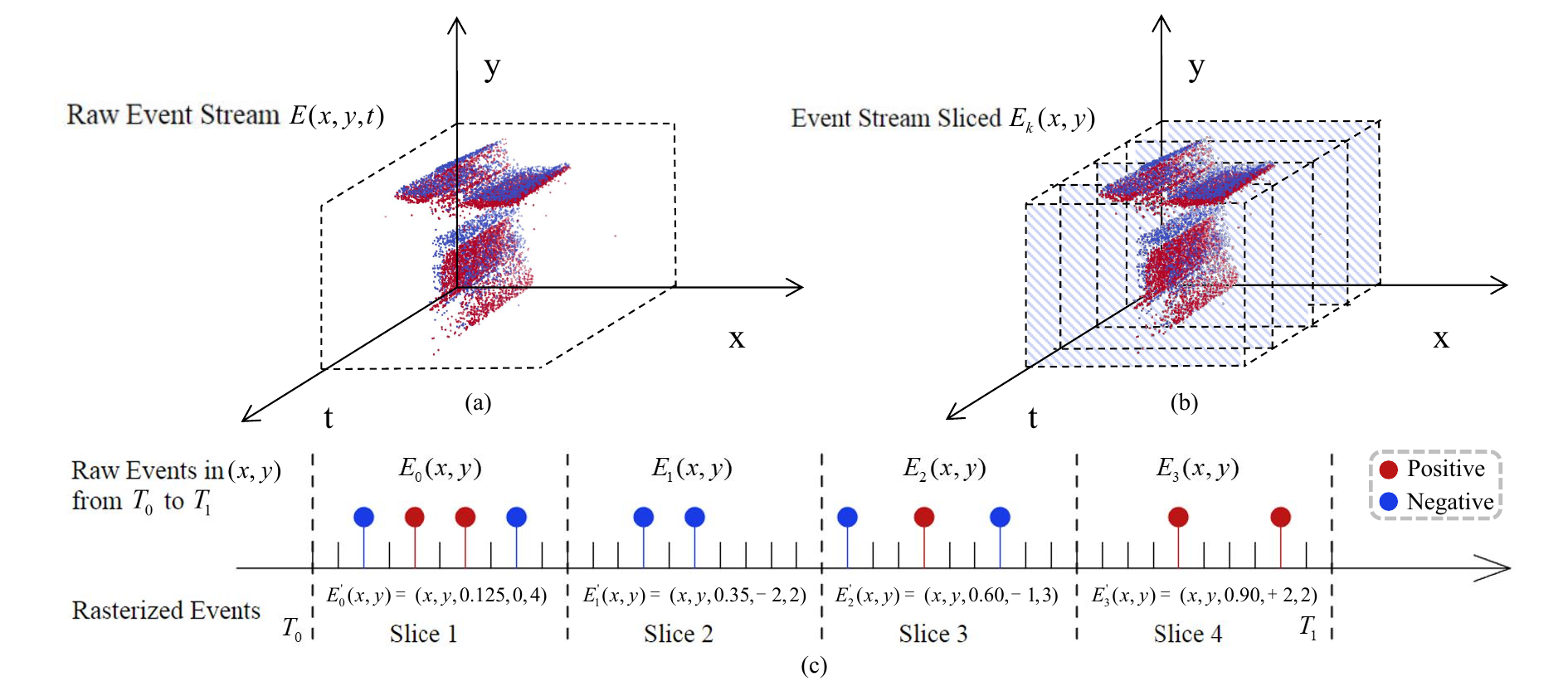} 
\caption{Schematic diagram of event point cloud rasterization: (a) Raw 3D event point cloud input, (b) Event stream sliced in the time dimension, and (c) Rasterized event point cloud at $(x, y)$ position. Note that the rasterization process preserves the discrete nature of the point cloud, rather than the 2D image.}
\label{ras}
\end{figure*}

\subsection{Overview}
In this section, we explore representing events in \([x,y,t]\) space using point cloud and decoupled voxel representations, offering an alternative to the dominating event frame design.
Datasets used in our study are first introduced (Sec.~\ref{sec:approach_dataset}) to illustrate how event streams are stored for subsequent 3D characterization. 
We then provide a detailed explanation of integrating the Rasterized Event Point Cloud (RasEPC) representation with existing point cloud backbone networks (see Sec.~\ref{sec:approach_EPC}), along with a comprehensive description of how the Decoupled Event Voxel (DEV) representation is implemented and utilized (see Sec.~\ref{sec:approach_DEV}).

\subsection{Dataset}
\label{sec:approach_dataset}
\noindent\textbf{DHP19 Dataset.} 
The DHP19 dataset~\citep{calabrese2019dhp19} is recorded by four synchronized DVS cameras and contains $33$ recordings of $17$ subjects of different sexes, ages, and sizes with $13$ joint annotations.
To preserve the 3D event information, we perform the denoising and filtering operations outlined in DHP19 on raw event point sets.
Each event point is represented by $e {=} (x, y, t, p)$, 
where $(x,y)$ is the pixel location of the event, $t$ represents the timestamp in microsecond resolution, and $p$ is the polarity of the brightness change, $0$ for decrease and $1$ for increase.  
To ensure data quality, we ignore the training data with points fewer than $1024$. 
Since the time resolution of the output label from the Vicon system is much lower than that of the event camera, we explore the relationship between events and labels. 
The \emph{Mean Label} setting adopted in the raw DHP19 dataset considers all views collectively.
When the total number of events from all four cameras reaches a predefined threshold, the label is generated by the mean value of all labels in the window. 
Specifically, if the threshold is set as $N{=}30k$, given by $E$ in Eq.~\ref{eq:1}, one 3D label, $gt_{mean}$ is produced.
The label is the mean value of the 3D coordinates for each joint generated in the window of $N$ events, followed by Eq.~\ref{eq:2}, and is shared by the four cameras.
\begin{equation}\label{eq:1}
E=\left\{E_{i} \mid i=1,2, \ldots N\right\},
\end{equation}
\begin{equation}\label{eq:2}
gt_{\text{mean}}=\operatorname{Mean}\left(gt_{T_{\min}}, gt_{T_{\min} +dt}, \ldots gt_{T_{\max}}\right),
\end{equation}
\begin{equation}
T_{\min } = \left\{\mathop{\arg\min}\limits_{T}\left(T-E_{1}(t)\right) \mid T \geq E_{1}(t) \right\}, 
\end{equation}
\begin{equation}
T_{\max } = \left\{\mathop{\arg\min}\limits_{T}\left(E_{N}(t)-T\right) \mid T \leq E_{N}(t)\right\}.
\end{equation}
After obtaining the 3D joint labels, we project them to 2D labels for a single camera view with the projection matrices.
We further explore the \emph{Last Label} setting, where a fixed number of events is counted for each camera view, and the label nearest to the last event is designated as the label for that particular camera. 
This setting leads to varying time spans of events across different cameras, and in Sec.~\ref{sec:experiment_ablation}, we compare the performance of these two settings.
Nevertheless, the DHP19 dataset is collected indoors and the range of activities captured in the recordings is narrow, many of which, such as leg kicking and arm abductions, do not often occur in real life. 
These gaps limit its application in real scenarios.

\noindent\textbf{MMHPSD Dataset.}
To evaluate more challenging scenarios, we also use the real-world MMHPSD dataset~\citep{zou2021eventhpe}.
This dataset comprises $240K$ frames, each containing synchronized images from an event camera, a polarization camera, and five RGB-D cameras.
It captures $15$ subjects performing $21$ distinct actions at three different speeds (fast, medium, slow).
The event camera features a resolution of $1280{\times}800$, and each subject contributes approximately $180$ video clips, totaling about $1.5$ hours of footage.
The dataset utilizes OpenPose~\citep{cao2017realtime} to detect 2D joints in all RGB images.

\noindent\textbf{EV-3DPW Dataset.} 
As previously discussed, the two real-world event datasets mentioned are limited to indoor scenarios with static backgrounds, restricted movement and activities, and no environmental occlusions.
To address these limitations, we introduce a new simulated dataset derived from the 3DPW dataset~\citep{von2018recovering} for investigating in-the-wild event HPE.
The 3DPW dataset is a challenging multiple-person dataset consisting of the training set of $22K$ images and the test set of $35K$ images, depicting various activities such as walking in the city, going upstairs, having coffee, or taking the bus. 
We crop the samples to segment each person into a distinct instance, which allows for seamless integration with detection algorithms, therefore enabling top-down multi-person HPE.
Each sequence is captured from a handheld mobile phone (resolution $1080{\times}1920$ pixels or $1920{\times}1080$ pixels; frequency $30Hz$), and 2D pose annotations are offered at a frequency of $30$ Hz, which matches the frame rate of the sequences.
We transform this dataset to events with a resolution similar to the widely used event camera DAVIS-346 while ensuring there is no introduction of horizontal or vertical proportional distortion: $256{\times}480$ pixels or $480{\times}256$ pixels.
We use the open-source simulator ESIM~\citep{rebecq2018esim_simulator} to convert videos into events.
To make the dataset closer to reality, DVS-Voltmeter~\citep{lin2022dvsvoltmeter} is employed to provide a noisy version.
The established public EV-3DPW dataset offers both noise-free and noisy versions.
Following the original 3DPW dataset, we exclusively utilize data that meets the criteria of having at least $6$ joints correctly detected out of the $18$ joint annotations.
We will release the conversion code in our implementation.

\begin{figure*}[!t]
\centering
\includegraphics[width=1.0\linewidth]{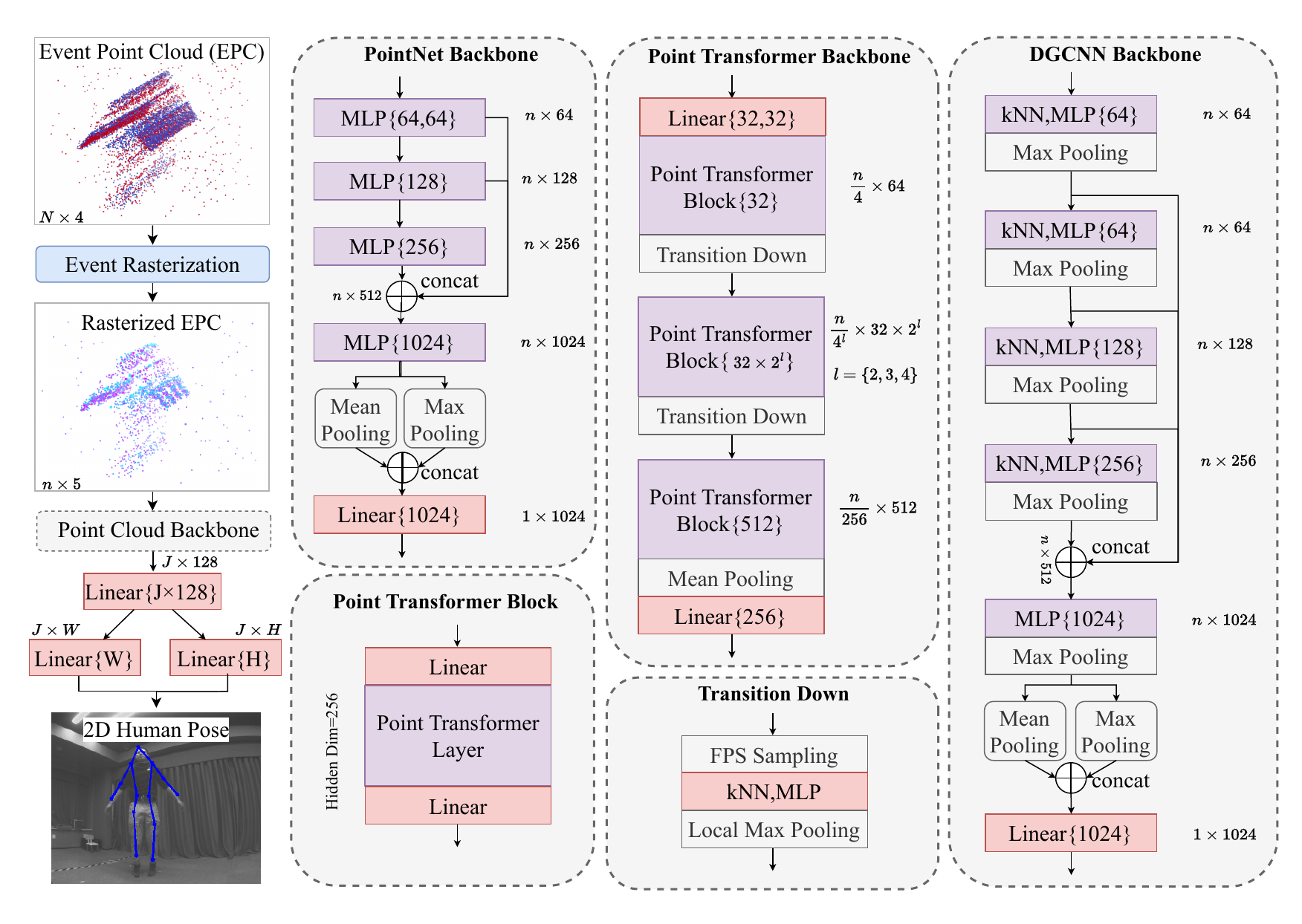}
\caption{
The proposed RasEPC pipeline.
The raw 3D event point cloud is first rasterized and then processed by the point cloud backbone. 
The features output by the backbone are then fed into linear layers to predict two vectors, which are decoded to obtain the 2D positions of human key points.}

\label{EPCbackbone}
\end{figure*}

\noindent\textbf{EV-JAAD Dataset.} 
To further evaluate the model's generalization capabilities when trained on the synthetic event dataset and applied to unseen scenes, we employ the identical data conversion procedures utilized for EV-3DPW to convert the JAAD dataset~\citep{rasouli2017they}.
JAAD comprises $346$ video clips with durations ranging from $5$ to $15$ seconds. 
These clips are collected in North America and Europe using a monocular camera with a resolution of $1920{\times}1080$ pixels.
The camera is positioned inside the car and below the rear-view mirror, running at $30FPS$, and human bounding boxes are provided.

\subsection{HPE based on 3D Rasterized Event Point Cloud}
\label{sec:approach_EPC}
\noindent\textbf{Rasterized Event Point Cloud Representation.} 
Event cameras offer microsecond-level time resolution, but using all recorded events for training can result in slow forward propagation and high memory consumption.
To address this problem, we propose an event rasterization method that aims to significantly reduce the number of events while retaining crucial information. 
The sparse representation of events after rasterization is termed the Rasterized Event Point Cloud (RasEPC).

In the event rasterization process, events between $t_i$ and $t_{i+1}$ are initially partitioned into $K$ time slices.
Within each small event slice, we condense events on each pixel to form a rasterized event.
Specifically, given all $M$ events on position $(x, y)$ in time slice $k$, where $p_{i}$ is converted to ${-}1$ for brightness decrease: 
\begin{equation}
E_k(x, y)=(x, y, t_i, p_i),\quad i=1,.., M.
\end{equation}
Then, we use the following equations to obtain the rasterized event $E^\prime_k(x,y)$:
\begin{equation}
E^\prime_k(x,y)=(x, y, t_{avg}, p_{acc}, e_{cnt}),
\end{equation}
\begin{equation}
t_{avg} = \frac{1}{M}\sum_i^Mt_i,\quad p_{acc} = \sum_i^Mp_i,\quad e_{cnt}=M.
\end{equation}
$K$ is selected as $4$ in this work to maintain the advantage of high-time-resolution. 
And $t_{avg}$ in all $K$ slices are normalized to the range of $[0, 1]$. 
An example of the proposed event point cloud rasterization is illustrated in Fig.~\ref{ras} and the result is visualized in Fig.~\ref{EPCbackbone}, where the color represents the value of $p_{acc}$ and the point size for $e_{cnt}$. 
Rasterization can be regarded as an online downsampling process, resulting in different numbers of events in different rasterized event point clouds. 
Since the number of input points in the point cloud processing network impacts both speed and accuracy, we further investigate the impact of the number of sampling points on the trade-off between efficiency and accuracy in human pose estimation in Sec.~\ref{sec:experiment_ablation}.

\noindent\textbf{Label Representation.}
In previous works~\citep{wei2016convposemachain, xiao2018simbase} based on RGB images with CNN models, estimating intermediate heatmaps of probabilities for each joint has generally been preferred over direct representation.
To make it applicable to event point clouds, we adopt a new coordinate representation called SimDR~\citep{li20212dsimdr}.
This approach converts the 2D labels $(x^{\prime}, y^{\prime})$ into two 1D heat-vectors, denoted as $\boldsymbol{p_{v}}$, which correspond to the size of the sensor.

These vectors are one-hot encoded and then further blurred using a Gaussian kernel, resulting in shapes of $(H, 1)$ and $(W, 1)$, respectively:
\begin{equation}\label{eq:8}
\left\{\begin{array}{c}
\begin{aligned}
&\boldsymbol{{p}_{v}} =\left[v_{0}, v_{1}, \ldots, v_{S}\right] \in \mathbb{R}^{S},\\
&v_{i} = \frac{1}{\sqrt{2 \pi} \sigma} \exp \left(-\frac{\left(i-v^{\prime}\right)^{2}}{2 \sigma^{2}}\right), 
\end{aligned}
\end{array}\right.
\end{equation}
where $v {\in} \left\{x, y\right\}$, $S \vert _{x} {=} W$, and $S \vert _{y} {=} H$.
We set ${\sigma}{=}8$ in our experiments as it proved to be the optimal choice for this task. 
Next, we normalize these vectors to have a maximum value of $1$ using Min-Max Normalization, resulting in the predicted vectors $\boldsymbol{\hat{p_{v}}}$. 
To determine the joint coordinates $(\hat{x}, \hat{y})$, we then perform the $argmax$ operation on these vectors to find the positions of the maximum values for the $x$ and $y$ axes, respectively:
\begin{equation}
\boldsymbol{{pred}_{v}} =\mathop{\arg\max}\limits_{j}\left(\boldsymbol{\hat{p_{v}}}(j)\right).
\end{equation}

\begin{figure*}[!t]
\centering
\includegraphics[width=1.0\linewidth]{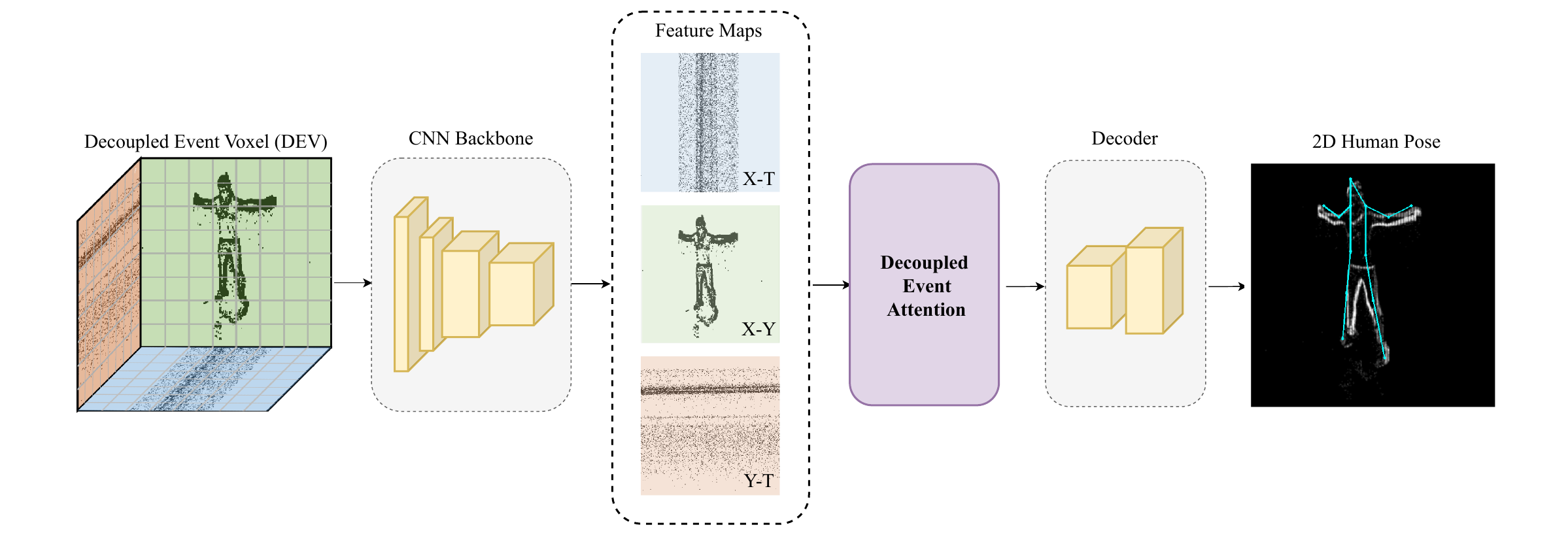}
\vskip-1.5ex
\caption{The proposed 3D decoupled event voxel pipeline. First, we discretize events into voxels and project them into three orthogonal directions. Features are extracted and reintegrated by combining 2D CNN and decoupled event attention. The keypoint heatmap is regressed through the decoder.} %
\label{DEVbackbone}
\end{figure*}

\noindent\textbf{3D Learning Model.}
As shown in Fig.~\ref{EPCbackbone}, the event point cloud is aggregated to a rasterized event point cloud, and features are extracted through the point cloud backbone.
These features are then processed through two linear layers to produce 1D vectors. 
We decode these $1$D vectors for the $x$-axis and $y$-axis separately to obtain the predicted joint locations.
In the field of LiDAR point cloud classification and segmentation, PointNet~\citep{qi2017pointnet}, DGCNN~\citep{wang2019dgcnn}, and Point Transformer~\citep{zhao2021pointtrans} are three common backbone networks, here we only modify and apply the encoder part to adapt to our task. 
For the PointNet backbone~\citep{qi2017pointnet}, we choose a $4$-layer MLP structure.
By incorporating multi-scale feature aggregation, we capture non-local information and subsequently obtain global features by max pooling and average pooling.
Considering that the event point clouds naturally have timing order, the joint alignment network is removed when using PointNet~\citep{qi2017pointnet}. 
For the other two backbone networks, we retain their original structures to verify the feasibility of utilizing event point clouds.
Our methods can be readily deployed and integrated with diverse 3D learning architectures and are well-suited for event-based HPE.

\subsection{HPE based on 3D Decoupled Event Voxel}
\label{sec:approach_DEV}
\noindent\textbf{Decoupled Event Voxel Representation.} 
In the field of LiDAR perception, voxel representations~\citep{wu20153d, maturana2015voxnet, qi2016volumetric} are often used.
Similarly, we can describe a 3D spatial-temporal scene with dense cubic features ${V}{\in}{\mathbb{R}^{{C}\times{H}\times{W}\times{T}}}$ where $H$, $W$, and $T$ are the resolution of the voxel space and $C$ denotes the feature dimension. 
Note that $T$ is set to be the same as $H$ and $W$.
A random point located at $(x, y, {\tau})$ in the real world can map to its voxel coordinates $(h, w, t)$ through one-to-one quantified correspondence $P_{vox}$, and the resulting feature $f_{x,y,{\tau}}$ can be described by sampling $V$ at the integer coordinate $(h, w, t)$:
\begin{equation}
    \begin{aligned}
        f_{x,y,{\tau}} &= S(V, (h, w, t)) \\
                                     &= S(V, P_{vox}(x,y,{\tau})),
    \end{aligned}
\end{equation}
where $S(value, position)$ denotes sampling from the $value$ at the specific $position$, resulting in storage and computation complexity of voxel features proportional to $O(HWT)$.
To make real-time prediction possible, we place the event stream into a voxel grid and project it to three orthogonal planes: $xy$, $x{\tau}$, and $y{\tau}$, which are described as \textbf{D}ecoupled \textbf{E}vent \textbf{V}oxel (DEV) planes:
\begin{equation}
    \begin{aligned}
        & D = [D^{HW}, D^{TH}, D^{WT}], \\
        & D^{HW} \in {\mathbb{R}^{{C}\times{H}\times{W}}},\\
        & D^{TH} \in {\mathbb{R}^{{C}\times{T}\times{H}}},\\
        & D^{WT} \in {\mathbb{R}^{{C}\times{W}\times{T}}}.
    \end{aligned}
\end{equation}
Given a query point at $(x, y, \tau)$ in the event stream, DEV representation tries to aggregate its projections on the tri-perspective views to obtain a spatiotemporal description of the point.
To elaborate, we project the point onto three planes to obtain the coordinates $[(h, w), (t, h), (w, t)]$, sample the DEV planes at these locations to retrieve the corresponding features $[f_{hw}, f_{th}, f_{wt}]$, and aggregate the three features to generate the final 
3D-aware feature $f_{DEV}$:
\begin{equation}
    \begin{aligned}
        & f_{hw} = S(D^{HW}, (h, w)) = S(D^{HW}, P_{hw}(x,y)),\\
        & f_{th} = S(D^{TH}, (t, h)) = S(D^{TH}, P_{th}({\tau},x)),\\
        & f_{wt} = S(D^{WT}, (w, t)) = S(D^{WT}, P_{wt}(y,{\tau})),\\
    \end{aligned}
\end{equation}
\begin{equation}
    f_{DEV} = \mathcal{A}(f_{hw},f_{th},f_{wt}),\\
\end{equation}
where the sampling function $S$ is implemented by nearest neighbor interpolation.
The aggregation function $\mathcal{A}$ is achieved by our newly proposed decoupled event attention module.
DEV representation creates a complete 3D feature space that resembles the voxel feature space but with significantly lower storage and computational complexity, specifically $O(HW{+}TH{+}WT)$. 
This complexity is one order of magnitude lower than that of the voxel approach.

\begin{figure*}[!t]
\centering
\includegraphics[width=1.0\linewidth]{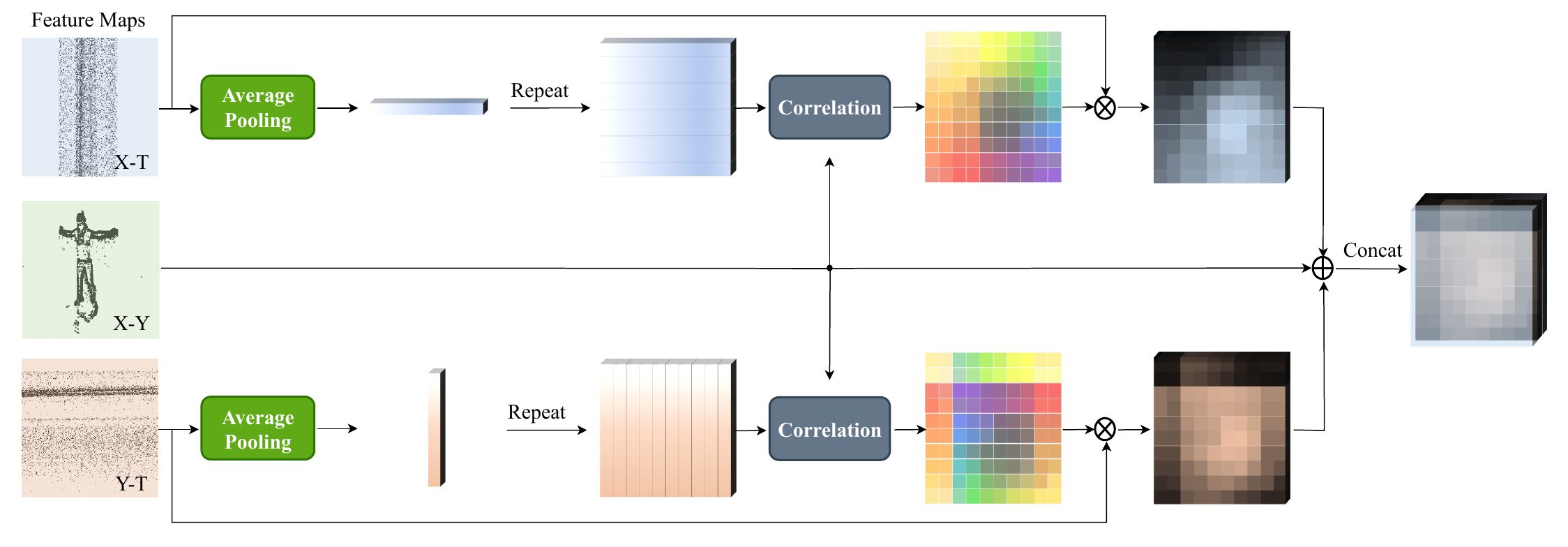}
\caption{Illustration of the proposed Decoupled Event Attention (DEA) module. Aggregate and copy the $xt$ and $yt$ plane features along the $t$-axis direction, respectively. The features of the two planes corresponding to the 2D coordinate plane are obtained by calculating their correlation with the $xy$ plane features, giving the per-pixel matching cost. 3D feature descriptions of the 2D coordinate plane are obtained via concatenation.}
\label{DEAModule}
\end{figure*}

\noindent\textbf{Network Architecture.}
The schematic diagram of the HPE method based on 3D decoupled event voxels is shown in Fig.~\ref{DEVbackbone}. 
Projections from three decoupled directions are fed into a CNN backbone network to extract multi-scale features for three views.
We employ three famous backbone networks for human pose estimation: DHP19~\citep{calabrese2019dhp19}, ResNet~\citep{xiao2018simbase}, and MobileHP~\citep{choi2021mobilehumanpose}. 
After extracting features, we aggregate them on the 2D $xy$ plane through the Decoupled Event Attention (DEA) module and use deconvolution to regress keypoints. 
Labels are processed as the common-practice 2D heatmaps $\boldsymbol{p}$:
\begin{equation}
\boldsymbol{p} = \exp \left(-\frac{\left(\boldsymbol{v}-\boldsymbol{v^{\prime}}\right)^{2}}{2 \sigma^{2}}\right),
\end{equation}
where $\boldsymbol{v}$ is a pixel of the heatmap in the position $(x, y)$, $\boldsymbol{v^{\prime}}$ is the target joint location. 
Gaussian kernel size ${\sigma}$ is related to the radius of the joint heatmap. 
We empirically set ${\sigma} {=} 2$ and heatmap size of $64{\times}64$. 
For a fair comparison, the heatmap size and Gaussian kernel size are identical for the 2D frame-based approach and our DEV method.

\noindent\textbf{Decoupled Event Attention Module.}
To establish the connection between the 2D $xy$ coordinate plane and the temporal-related $x{\tau}$ and $y{\tau}$ planes respectively, we introduce the \textbf{D}ecoupled \textbf{E}vent \textbf{A}ttention (DEA) module. 
The structure of the module is shown in Fig.~\ref{DEAModule}. 
The DEA module's design is inspired by using the $xy$ coordinate plane as an index plane, finding the most likely corresponding $y(x)$ value for each $x(y)$ in the $x\tau(y\tau)$ plane through correlation, thus retrieving 3D cues from the tri-view DEV representation.
For the features of the $x{\tau}$ and $y{\tau}$ planes, they are aggregated along the ${\tau}$ direction to obtain global features in the $x$-axis or $y$-axis direction. 
Subsequently, the global features are expanded to match the size of the two-dimensional coordinate plane, and correlations with the $xy$ plane are established.
The step of calculating the correlation between the aggregated features and the feature in the $xy$ plane can be seen as the process of retrieving 3D cues from three projection planes. 
Adding the ${\tau}$-related features reasonably and effectively to the $xy$ plane can take the temporal information into consideration, thereby improving estimation accuracy.

Specifically, given three projection planes $D^{HW}$, $D^{TH}$, $D^{WT}$, the feature maps ${f_{hw}}\in{\mathbb{R}^{{C}\times{H}\times{W}}}$, ${f_{th}}\in{\mathbb{R}^{{C}\times{T}\times{H}}}$, ${f_{wt}}\in{\mathbb{R}^{{C}\times{W}\times{T}}}$ are obtained by the encoder $e(\cdot)$. 
We first aggregate the global context in the $x$ and $y$ directions respectively and expand the features to the size of (${C}\times{H}\times{W}$) by repeating them $W$ or $H$ times:
\begin{equation}\label{eq:11}
\left\{\begin{array}{c}
\begin{aligned}
&{\Hat{f_{th}}(i, j)} = Repeat_T(\frac{1}{T} \sum_{k=1}^T {f_{th}(i, k ,j)}, W),\\
&{\Hat{f_{wt}}(i, j)} = Repeat_T(\frac{1}{T} \sum_{k=1}^T {f_{wt}(i, j, k)}, 
H),\end{aligned}
\end{array}\right.
\end{equation}
where the average pooling window is ${T}{\times}{1}$ and ${1}{\times}{T}$, respectively. 
$Repeat_{T}(x, n)$ indicates replicating the feature $x$ along the $T$ dimension $n$ times.
After that, we perform dot product between the extended global features ($\Hat{f_{th}}(i, j)$ and $\Hat{f_{wt}}(i, j)$) and the $xy$ plane feature to obtain the $x$-axis and $y$-axis correlation matrix ${C_h, C_w} {\in} \mathbb{R}^{{H} \times {W}}$, which encode the non-local visual similarity:
\begin{equation}\label{eq:12}
\left\{\begin{array}{c}
\begin{aligned}
&{C_h(h, w)} = {\Hat{f_{th}}} \cdot {f_{hw}},\\
&{C_w(h, w)} = {\Hat{f_{wt}}} \cdot {f_{hw}}.
\end{aligned}
\end{array}\right.
\end{equation}
The features of $xt$ and $yt$ plane ${f_{th}}$, ${f_{wt}}$ are multiplied by the corresponding correlation matrix ${C_h, C_w}$, and then we concatenate them with ${f_{hw}}$ to obtain a 2D feature map carrying the 3D information of spatial-temporal space:
\begin{equation}\label{eq:13}
 f_{DEV} = f_{hw}~\oplus~({C_h}\cdot{f_{th}})~\oplus~({C_w}\cdot{f_{wt}}),  \\
\end{equation}
where ${\oplus}$ represents the concatenation operation.
Note that we set time bins $T$ identical to $H$ and $W$ to achieve this retrieval process. 
The intuition of the operation lies in that it emphasizes global features along the $x$-axis and $y$-axis, which establishes correspondence with points on the $xy$ plane through correlation.

\section{Experiments}
\subsection{Experiment Setups}
To verify the proposed 3D event representations, we conduct experiments on the indoor single-person DHP19 dataset~\citep{calabrese2019dhp19}, the challenging MMHPSD dataset, and our synthetic outdoor multi-person EV-3DPW dataset.
The DHP19 dataset~\citep{calabrese2019dhp19} follows the original split setting, using $S1{\sim}S12$ for training and 
$S13{\sim}S17$
for testing. 
When estimating 3D keypoint positions, only two front cameras are used following the raw DHP19 dataset. 
The MMHPSD dataset~\citep{zou2021eventhpe} uses subjects $01$, $02$, and $07$ for testing and the remaining $13$ subjects for training, consistent with the original dataset's protocol.
In the simulated EV-3DPW dataset, there are $23,475$ train samples and $40,145$ test samples according to the split standard of the 3DPW dataset~\citep{von2018recovering}. 
We train our models and test the speed of inference with a batch size of $1$ on a single RTX 3090 GPU, implemented in PyTorch.
For HPE based on the Rasterized Event Point Cloud (RasEPC), 
we employ the Adam optimizer~\citep{kingma2014adam} for $30$ epochs with an initial learning rate of $1e^{-4}$ dropping by $10$ times at $15$ and $20$ epoch. 
The Kullback–Leibler divergence loss is used for supervision. 
To simulate real-time estimation application scenarios, we further evaluate our approach on an NVIDIA Jetson Xavier NX edge computing platform. 
For HPE based on decoupled event voxel (DEV), we employ the Adam optimizer~\citep{kingma2014adam} with a learning rate of $1e^{-3}$ for two real-world datasets and $1e^{-4}$ for the outdoor dataset EV-3DPW to optimize our network.
The batch size is set to $32$ with MSE Loss as the loss function. 
The selection of hyperparameters and loss functions for the event frame and RGB image methods is consistent with the DEV method. 
We evaluate the results through the Mean Per Joint Position Error (MPJPE), commonly used in human pose estimation:
\begin{equation}\label{eq:14}
\mathrm{MPJPE}=\frac{1}{J} \sum_{i}^{J}\left\| pred_{i}-gt_{i} \right\|_{2},
\end{equation}
which equals the average Euclidean distance between the ground truth and prediction. 
The metric space for 2D error is pixels, and millimeters for 3D error.

\subsection{Ablation Studies}
\label{sec:experiment_ablation}
\noindent\textbf{Experiments based on Rasterized Event Point Cloud (RasEPC).}
To demonstrate the superiority of rasterizing event point clouds over raw point cloud input, we conduct ablation experiments on the DHP19 dataset using the well-known PointNet backbone with $2048$ points, as shown in Table~\ref{tab:rast}.
The timestamps of the raw event data are in microseconds, resulting in a large scale difference from the x-y axis, so the performance of raw input is unsatisfactory. 
Normalizing the timestamp into the range $[0, 1]$ and changing the polarity of $0$ to ${-}1$ can reach better results. 
When combined with rasterization, MPJPE further decreases. 
We take all the five-channel representations $(x, y, t_{avg}, p_{acc}, e_{cnt})$ as the best choice. 
As a necessary preprocessing step, the event point cloud rasterization can be easily employed in real-time applications by leveraging a buffer to rapidly update event information in all channels with a preset time window length.
RasEPC preserves the high temporal resolution of event cameras and is ideal for real-time processing. 

\begin{table}[t]
   \caption{\textbf{Event point cloud rasterization ablations.}}
   \label{tab:rast}
   \centering
   \begin{adjustbox}{width=0.5\textwidth,center}
   \renewcommand\arraystretch{1.2}{\setlength{\tabcolsep}{8mm}{\begin{tabular}{ccccc} 
            \toprule
            Input & Channel & MPJPE$_{2D}$ & MPJPE$_{3D}$   
            \\
            
            \midrule
            
            \multirow{2}{*}{Raw} 
            & $x,y,t$ & 24.75 & 310.65  \\
            & $x,y,t,p$ & 24.74 & 310.64  \\
            
            \midrule
            
            \multirow{2}{*}{Normalized} 
            & $x,y,t_{norm}$ & 7.92 & 89.62  \\
            & $x,y,t_{norm},p_{\pm1}$ & 7.61 & 86.07  \\
            
            \midrule
            
            \multirow{3}{*}{Rasterized} 
            & $x,y,t_{avg}$ & 7.77 & 87.59  \\
            & $x,y,t_{avg},p_{acc}$ & 7.40 & 84.58  \\
            & $x,y,t_{avg},p_{acc},e_{cnt}$ & \textbf{7.29} & \textbf{82.46}  \\

            \bottomrule
         \end{tabular}}}
     \end{adjustbox}
\end{table}

\begin{table}[!t]
\caption{\textbf{Event point cloud ablation experiment on time slice $K$.}}
   \begin{adjustbox}{width=0.5\textwidth,center}
   \renewcommand\arraystretch{1.2}{\setlength{\tabcolsep}{8mm}{\begin{tabular}{ccccc} 
            \toprule
            \multirow{2}{*}{$K$} & \multicolumn{2}{c}{\underline{\quad PointNet-2048 \quad}} & \multicolumn{2}{c}{\underline{\quad PointNet-4096 \quad}}\\
            & MPJPE$_{2D}$ & MPJPE$_{3D}$ & MPJPE$_{2D}$ & MPJPE$_{3D}$ \\
            \midrule
            1 & 7.29 & 82.40 & 7.24 & 81.74 \\
            2 & \textbf{7.28} & 82.49 & 7.23 & 81.99 \\
            4 & 7.29 & \textbf{82.46} & \textbf{7.21} & \textbf{81.42} \\
            8 & 7.31 & 83.18 & 7.23 & 81.67 \\
            
            \bottomrule
         \end{tabular}}}
     \end{adjustbox}
\label{tab:K_ablation}
\end{table}

We additionally conduct an ablation study on time slice $K$ used in rasterization shown in Table~\ref{tab:K_ablation}, which has an impact on both information density and time resolution. 
When it is smaller, the information density is higher but the time resolution is lower, thus the choice needs to be weighed. 
Since the HPE task is not sensitive to time slice and achieves satisfactory performance at $4$, we select $K{=}4$ in our task.

In a point cloud backbone network, the number of input points impacts both the model's speed and accuracy.
As shown in Table~\ref{tab:number}, we test PointNet on the DHP19 dataset using the rasterized event point format on an NVIDIA Jetson Xavier NX edge computing platform.
In general, more points lead to higher accuracy accompanied by a decrease in speed. 
When it reaches $7500$ points, the accuracy starts to decline due to the sampling strategy with replacement, \textit{i.e.}, repeated sampling when the number of points is insufficient. 
To hold a fine trade-off between speed and accuracy, we choose $2048$ points for other experiments.

As mentioned in Sec.~\ref{sec:approach_dataset}, we have introduced two label settings for event cameras: \emph{Mean Label} and \emph{Last Label}, given the significantly lower temporal resolution of the output labels from the Vicon system compared to event cameras.
Rather than assigning labels for an instant, we aim to establish labels within a tiny time window.
Here, we test two settings on all three backbones in Table~\ref{tab:label}. 
From a theoretical perspective, the \emph{Last Label} setting seems more reasonable for real-time prediction. 
But the \emph{Mean Label} setting performs better in our task, which is attributed to the large temporal span between early events and labeled time points, as well as the single measurement error in instantaneous label recording.

\begin{table}[t]
   \caption{\textbf{Ablations on the number of sampling points.}}
   \label{tab:number}
   \centering
   \begin{adjustbox}{width=0.5\textwidth,center}
   \renewcommand\arraystretch{1.2}{\setlength{\tabcolsep}{8mm}{\begin{tabular}{cccc}
            \toprule
            Sampling Number & MPJPE$_{2D}$ & MPJPE$_{3D}$ & Latency (ms)   
            \\
            
            \midrule
            
            1024 & 7.49 & 85.14 & \textbf{9.43}  \\
            2048 & 7.29 & 82.46 & 12.29  \\
            4096 & \textbf{7.21} & \textbf{81.42} & 18.80  \\
            7500 & 7.24 & 81.72 & 29.18  \\

            \bottomrule
         \end{tabular}}}
    \end{adjustbox}
\end{table}

\begin{table}[!t]
   \caption{\textbf{Comparison of different point cloud labels.}}
   \label{tab:label}
   \centering
   \begin{adjustbox}{width=0.5\textwidth,center}
   \renewcommand\arraystretch{1.2}{\setlength{\tabcolsep}{8mm}{\begin{tabular}{ccc} 
            \toprule
            \multirow{2}{*}{Method}&\multicolumn{2}{c}{\underline{\quad  \quad MPJPE$_{2D}$  \quad \quad}} \\
            & Last Label & Mean Label\\ 
            
            \midrule
            
            PointNet~\citep{qi2017pointnet}                & 7.50 & \textbf{7.29}\\
            DGCNN~\citep{wang2019dgcnn}                   & 6.96 & \textbf{6.83}\\
            Point Transformer~\citep{zhao2021pointtrans}        & 6.74 & \textbf{6.46}\\
    
            \bottomrule
         \end{tabular}}}
     \end{adjustbox}
\end{table}

\noindent\textbf{Experiments based on Decoupled Event Voxel (DEV).} 
To evaluate the effectiveness of the proposed decoupled event voxel representation, we conduct ablation studies compared with the event frame approach based on two backbones on the DHP19 dataset. 
The results, presented in Table~\ref{tab:dev_framework}, demonstrate a substantial increase in accuracy for both backbones when utilizing DEV representation along with the DEA module.
The DHP19 backbone exhibits a more pronounced improvement, primarily due to its simple structure.

To investigate the impact of the two additional temporal-related perspectives introduced by the proposed DEV representation, we conduct experiments to assess the contributions of three orthogonal views using the DHP19 backbone on the DHP19 dataset in Table~\ref{tab:dev_views}.
Incorporating temporal information from either $x$ or $y$ direction, along with event frames, has led to improved accuracy.
Injecting information from the $xt$ plane leads to a lower MPJPE than the $yt$ plane.
We believe this is because most movements in the dataset involve changes in a person's body width ($x$-direction) rather than significant changes in their height ($y$-direction).
Information about $x$-direction helps the network better understand movements involving the expansion of arms or sidekicks, which can lead to significant displacements of keypoints in the lateral direction. 
Therefore, adding lateral information provides more contextual information about keypoint positions, thereby improving estimation accuracy.
Utilizing the information from the three projection planes further enhances the network's accuracy.

\begin{table}[!t]
\caption{\textbf{Ablations on 3D Decoupled Event Voxel Framework.}}
   \centering
   \begin{adjustbox}{width=0.5\textwidth,center}
   \renewcommand\arraystretch{1.2}{\setlength{\tabcolsep}{8mm}{\begin{tabular}{lcccc} 
            \toprule
            Method & DEV & MPJPE$_{2D}$ & MPJPE$_{3D}$ & Gain \\
            
            \midrule
            
            DHP19~\citep{calabrese2019dhp19} & \textit{w/o} & 7.67 & 87.90 & - \\
            DEV-Pose (DHP19) & \textit{with}  & \textbf{6.27} & \textbf{71.01}  & $\uparrow$18.25\%\\

            \midrule

            MobileHP-S†~\citep{choi2021mobilehumanpose} & \textit{w/o} & 5.65 & 64.14 & -\\
            DEV-Pose (MobileHP-S) & \textit{with} & \textbf{5.20} & \textbf{58.80} & $\uparrow$7.96\%\\
            
            \bottomrule
         \end{tabular}}}
    \end{adjustbox}
\label{tab:dev_framework}
\end{table}

\begin{table}[!t]
\caption{\textbf{Ablations on event views of DEV-Pose.}}
   \centering
   \begin{adjustbox}{width=0.5\textwidth,center}
   \renewcommand\arraystretch{1.2}{\setlength{\tabcolsep}{8mm}{\begin{tabular}{ccccc} 
            \toprule
            \multicolumn{3}{c}{Event Views} & \multirow{2}{*}{MPJPE$_{2D}$ }& \multirow{2}{*}{MPJPE$_{3D}$ }  \\
            \cline{1-3}
            $xy$ & $xt$ & $yt$ & & \\
            
            \midrule
            
            \checkmark & - & - & 7.67 & 87.90  \\
            \checkmark & \checkmark & - & 6.42 & 72.45  \\
            \checkmark & - & \checkmark & 6.60 & 74.90  \\
            \checkmark & \checkmark & \checkmark & \textbf{6.27} & \textbf{71.01}  \\

            \bottomrule
         \end{tabular}}}
    \end{adjustbox}
\label{tab:dev_views}
\end{table}

\begin{table}[!t]
\caption{\textbf{Ablations on view aggregation method.}}
   \centering
   \begin{adjustbox}{width=0.5\textwidth,center}
   \renewcommand\arraystretch{1.2}{\setlength{\tabcolsep}{8mm}{\begin{tabular}{lcccc} 
            \toprule
            Aggregation method & MPJPE$_{2D}$ & MPJPE$_{3D}$ \\
            
            \midrule
            $w/o$ & 5.37 & 61.03 \\
            Add & 5.14 & 58.09 \\
            Concat & 5.09 & 57.59 \\
            Attention Feature Fusion~\citep{dai2021attentional} & 5.07 & 57.29 \\
            Decoupled Event Attention (Ours) & \textbf{4.93} & \textbf{55.53} \\
            \bottomrule
         \end{tabular}}}
    \end{adjustbox}
\label{tab:dev_view_aggregation}
\end{table}

Moreover, we investigate the most effective feature fusion strategy in Table~\ref{tab:dev_view_aggregation}.
We conduct experiments using Pose-ResNet18† as the backbone, trained on the DHP19 dataset.
Regardless of the specific feature fusion strategy, the introduction of temporal-related information consistently improves accuracy.
We test three commonly used feature fusion methods, namely addition, concatenation, and Attention Feature Fusion~\citep{dai2021attentional}, compared against our DEA fusion method. 
Directly adding features from three perspectives into a single feature results in mixed information, which challenges the model's ability to discern which information is more important for the task, ultimately constraining its performance. 
Concatenation and Attention Feature Fusion preserve the distinctiveness of information from each perspective, empowering the model to independently leverage each source of information, resulting in improved performance.
Our novel Decoupled Event Attention (DEA) module, utilizing the spatial correspondence of three orthogonal planes, offers a more comprehensive depiction of points on a two-dimensional coordinate plane.
Among the above four feature fusion strategies, our approach performs the best, which leads to $8.2\%$ and $9.0\%$ error reductions in terms of MPJPE$_{2D}$ and MPJPE$_{3D}$, respectively, compared to the single-view baseline.

\begin{table}[!t]
   \caption{\textbf{Comparison of different pooling methods.}}
   \label{tab:dev_pool}
   \centering
   \begin{adjustbox}{width=0.5\textwidth,center}
   \renewcommand\arraystretch{1.2}{\setlength{\tabcolsep}{8mm}{\begin{tabular}{lccc} 
            \toprule
            \multirow{2}{*}{Method}&\multicolumn{2}{c}{\underline{\quad \quad  \quad MPJPE$_{2D}$  \quad \quad \quad}} \\
            & Max Pooling & Average Pooling\\ 
            
            \midrule
            
            DEV-Pose (DHP19)               & 6.38  & \textbf{6.27}\\      
            DEV-Pose (MobileHP-S)          & 5.37 & \textbf{5.20}\\
            DEV-Pose (ResNet18)            & 5.02 & \textbf{4.93}\\

            \bottomrule
         \end{tabular}}}
     \end{adjustbox}
\end{table}

\begin{table}[!t]
\caption{\textbf{Ablations on different event representations.}}
   \centering
   \begin{adjustbox}{width=0.5\textwidth,center}
   \renewcommand\arraystretch{1.2}
   \setlength{\tabcolsep}{8mm}
   \begin{tabular}{lcccc}
            \toprule
            Event Representation & MPJPE$_{2D}$ & MPJPE$_{3D}$ \\
            
            \midrule
            2D Event Frame & 7.67 & 87.90 \\
            HATS~\citep{sironi2018hats} & 7.15 & 81.13 \\
            Ev-FlowNet~\citep{zhu2018ev} & 6.78 & 77.66 \\
            EST~\citep{gehrig2019end} & 7.32 & 83.27 \\
            DEV-Pose (Ours) & \textbf{6.27}  & \textbf{71.01} \\
            \bottomrule
   \end{tabular}
   \end{adjustbox}
\label{tab:ablation_HATS_EVFlowNet_DEV_EST}
\end{table}

\begin{table*}[!t]
   %
   \caption{\textbf{3D human pose estimation on the DHP19 dataset.}}
   \label{tab:compare_DHP19}
   \centering
   \begin{adjustbox}{width=1\textwidth,center}
   \renewcommand\arraystretch{1.2}{\setlength{\tabcolsep}{8mm}{\begin{tabular}{clcccc}
            \toprule
            Input & Method & MPJPE$_{2D}$ & MPJPE$_{3D}$ & $\#$Params (M) & Latency (ms)   
            \\
            
            \midrule
            
            \multirow{4}{*}{2D Event Frames} 
            & DHP19~\citep{calabrese2019dhp19} & 7.67 & 87.90 & \textbf{0.22} & \textbf{1.80} \\
            & MobileHP-S†~\citep{choi2021mobilehumanpose} & 5.65 & 64.14 & 1.83 & 10.9 \\
            & Pose-ResNet18†~\citep{xiao2018simbase} & 5.37 & 61.03 & 15.4 & 6.14 \\
            & Pose-ResNet50†~\citep{xiao2018simbase} & \textbf{5.28} & \textbf{59.83} & 34.0 & 13.0 \\
  
            \midrule

            \multirow{3}{*}{\textbf{3D Event Point Cloud}}%
            & RasEPC (PointNet~\citep{qi2017pointnet}) & 7.29 & 82.46 & 4.46 & \textbf{4.48}\\
            & RasEPC (DGCNN~\citep{wang2019dgcnn}) & 6.83 & 77.32 & 4.51 & 11.3 \\
            & RasEPC (Point Transformer~\citep{zhao2021pointtrans}) & \textbf{6.46} & \textbf{73.37} & \textbf{3.65} & 161 \\ %
            
            \midrule
            
            \multirow{2}{*}{3D Voxel-based Methods}
            & VMV-PointTrans~\citep{xie2022vmv, zhao2021pointtrans} & 9.13 & 103.23 & 3.67 & - \\
            & VMST-Net~\citep{liu2023voxel} & \textbf{6.45} & \textbf{73.07} & \textbf{3.59} & - \\

            \midrule
            
            \multirow{2}{*}{\textbf{3D Decoupled Event Voxel}}
            & DEV-Pose (DHP19) & 6.27 & 71.01 & \textbf{0.91} & \textbf{3.92}\\       
            & DEV-Pose (ResNet18) & \textbf{4.93} & \textbf{55.53} & 23.7 & 12.0 \\    

            \bottomrule
         \end{tabular}}}
     \end{adjustbox}
\end{table*}

In Fig.~\ref{DEAModule}, we aggregate features from the $xt$ and $yt$ planes along the time ($t$) axis through pooling in the DEA module.
To explore which pooling method is more suitable for human pose estimation tasks, we conduct experiments with three commonly used backbones, applying both max pooling and average pooling.
The results presented in Table~\ref{tab:dev_pool} indicate that average pooling outperforms max pooling across all three backbones.
This superiority of average pooling is due to its robustness in the presence of noise interference, which is a common challenge when dealing with event-based information.
Therefore, average pooling is recommended for event information aggregation for the DEA module.

To further explore the effects of different event representations, we conduct ablation experiments on four event representation forms, namely, 2D event frames, HATS~\citep{sironi2018hats}, Ev-FlowNet~\citep{zhu2018ev}, EST~\citep{gehrig2019end}, and our DEV-Pose, on the DHP19 dataset based on the dhp19 backbone. 
The results are shown in Table \ref{tab:ablation_HATS_EVFlowNet_DEV_EST}.
Compared to 2D event frame methods, both HATS~\citep{sironi2018hats}, Ev-FlowNet~\citep{zhu2018ev}, and EST~\citep{gehrig2019end} enhance the accuracy of human pose estimation, with our DEV representation showing additional improvement.

\subsection{Comparison of Event Representations}

\begin{figure}[!t]
\centering
\hspace{-5mm} 
\includegraphics[width=0.95\linewidth]{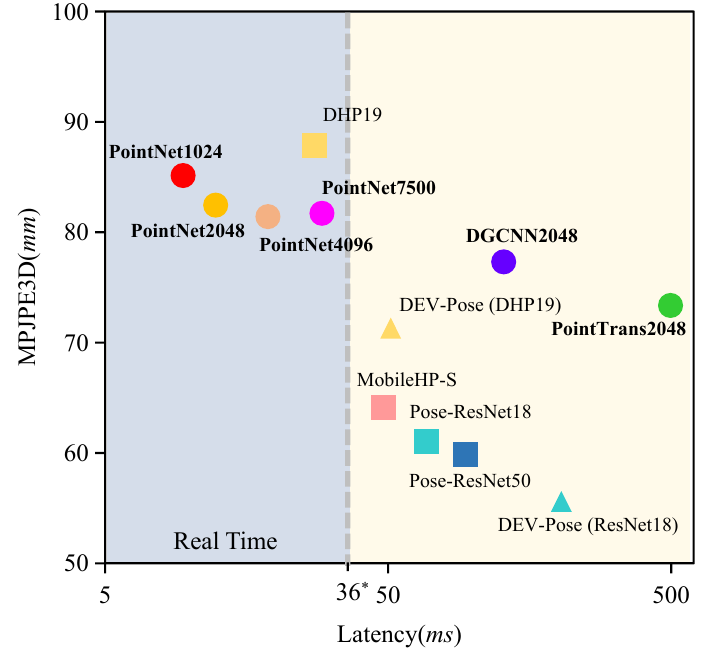}
\vskip-1ex
\caption{Latency \textit{vs.} Mean Per Joint Position Error with a logarithmic x-axis of 2D CNN backbones (square markers) our Event Point Cloud pipeline (circular markers), and our Decoupled Event Voxel pipeline (triangle markers). *The real-time criterion is statistically obtained on the DHP19 test dataset~\citep{calabrese2019dhp19}.
}
\label{EPC_speed}
\end{figure}

\noindent\textbf{Results on the DHP19 Dataset.}
In Table~\ref{tab:compare_DHP19}, we present a comparative analysis of four event representations: 2D event frames, our 3D rasterized event point clouds (RasEPC), 3D voxel-based methods and our 3D decoupled event voxels (DEV), all evaluated on the DHP19 dataset († indicates our reimplementation). 
We follow the Simple Baseline~\citep{xiao2018simbase} to train the models of Pose-ResNet18 and Pose-ResNet50 with constant count event frames. 
MobileHumanPose~\citep{choi2021mobilehumanpose} is tested as another backbone for 2D prediction with the same framework as ours. 
The event frame approach exhibits higher accuracy than the event point cloud approach, which is consistent with expectations, given the greater number of parameters in the latter. 
Although the speed advantage of the event point cloud is not pronounced on a single RTX 3090 GPU, it shines on an NVIDIA Jetson Xavier NX edge computing platform as shown in Fig.~\ref{EPC_speed}, where our RasEPC (PointNet~\citep{qi2017pointnet}) achieves the fastest inference with good performance.
Our PointNet only has a latency of $12.29ms$, which is ideally suitable for efficiency-critical scenarios.
Additionally, in the supplementary material, we provide qualitative results on the DHP19 dataset. 
Our findings reveal that the proposed event-point-cloud-based approach demonstrates increased robustness in managing static limbs compared to the original DHP19 model. 
This is particularly evident when limbs remain static during motion.
In these instances, a scarcity of generated events can cause certain parts to become invisible in the accumulated event frames. 
However, our RasEPC retains these sparse events, which may be a key factor in the improved accuracy of our method for predicting static limb keypoints.

\begin{figure*}[t]
\centering
\includegraphics[width=1.0\linewidth]{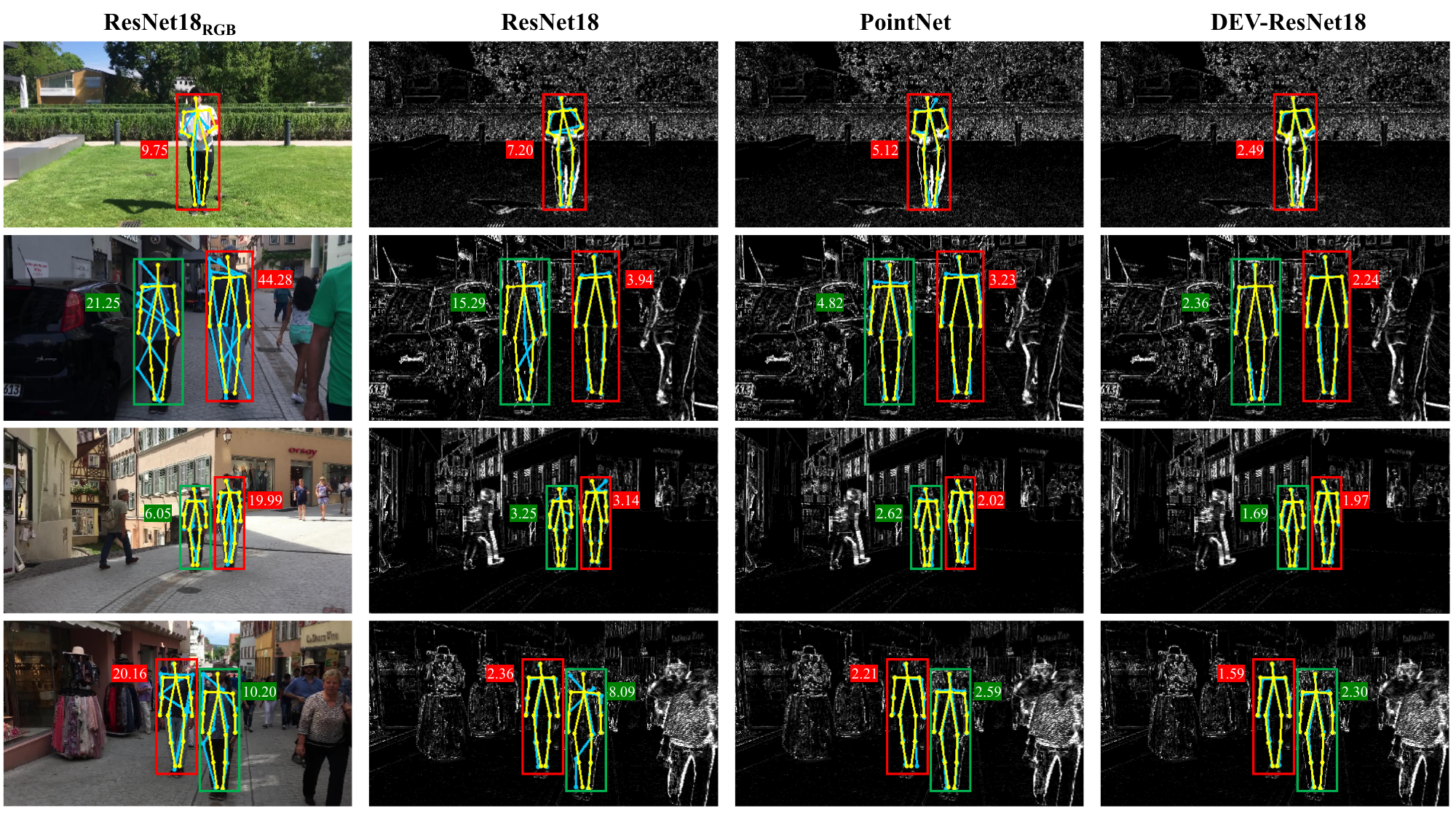}
\vskip-2ex
\caption{Results visualization for different models on the EV-3DPW dataset. The bounding box is obtained through the label, in which the result of a single person is shown (yellow for ground truth, blue for prediction). Compared to using RGB images, using events as input is easier to generalize to different scenarios. Our proposed two 3D event representations cope better with static limbs than processing events into event frames.}
\label{compare}
\vskip-2ex
\end{figure*}

\begin{figure*}[!t]
\centering
\includegraphics[width=1.0\linewidth]{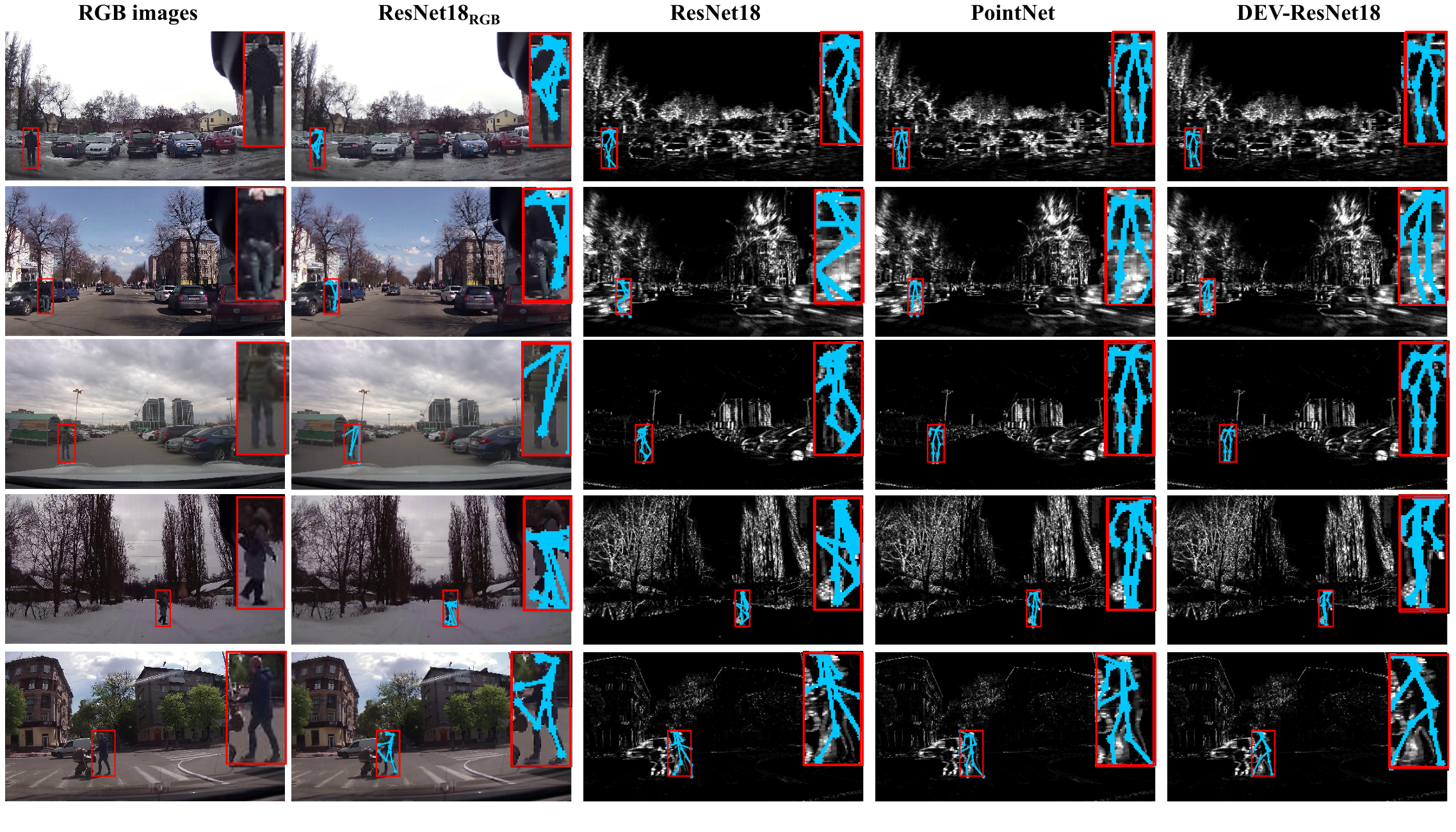}
\vskip-3ex
\caption{Qualitative results on the EV-JAAD dataset. The human bounding boxes are offered by the original JAAD dataset. Our two three-dimensional event representation methods successfully generalize from the synthetic dataset to unseen driving scenes.}
\label{compare_EVJAAD}
\end{figure*}

Compared to the event frame method under the same backbone, our decoupled event voxel method reduces the MPJPE value to a great extent without changing the parameter amount significantly. 
DEV-Pose (ResNet18) decreases MPJPE from $5.28$ to $4.93$ with much smaller parameters than Pose-ResNet50† with lower latency, which shows the effectiveness of adding temporal-related information.
Compared with the voxel-based method, MPJPE of DEV-Pose (DHP19) with a parameter size of $0.91M$ is better than the VMST-Net method with a parameter size of $3.59M$ ($6.27$ pixels \textit{vs.} $6.45$ pixels), demonstrating the effectiveness of DEV representation.
DEV-Pose (ResNet18) yields the highest accuracy while maintaining high efficiency ($12.0ms$), indicating that the proposed method achieves a better trade-off between accuracy and latency than the event frame method.

\noindent\textbf{Results on the MMHPSD Dataset.}
\begin{table}[!t]
   \caption{\textbf{Event-based HPE on the MMHPSD dataset.}}
   \label{tab: MMHPSD}
   \centering
   \begin{adjustbox}{width=0.5\textwidth}
   {
   {\begin{tabular}{clcccc}
        \toprule
        Input & Method & MPJPE$_{2D}$ 
        \\

        \midrule
        
        \multirow{6}{*}{2D Event Frames} 
        & DHP19†~\citep{calabrese2019dhp19} & 5.24  \\
        & MobileHP-S†~\citep{choi2021mobilehumanpose} & 3.93 \\
        & Pose-ResNet18†~\citep{xiao2018simbase} & 3.94 \\
        & Pose-ResNet50†~\citep{xiao2018simbase} &  \textbf{3.82} \\
        & SPHP (Conv.)~\citep{lin2023sparse} & 4.66 \\
        & SPHP (Sparse Conv.)~\citep{lin2023sparse} & 9.75 \\

        \midrule
        
        \multirow{3}{*}{3D Event Point Cloud}
        & RasEPC (PointNet~\citep{qi2017pointnet}) & 5.90  \\
        & RasEPC (DGCNN~\citep{wang2019dgcnn}) & 5.21  \\    
        & RasEPC (Point Transformer~\citep{zhao2021pointtrans}) & \textbf{4.79} \\       
        \midrule
        
        \multirow{2}{*}{3D Decoupled Event Voxel}
        & DEV-Pose (DHP19) & 3.28 \\      
        & DEV-Pose (ResNet18) & \textbf{2.53}  \\ 

        \bottomrule
        \end{tabular}}}
    \end{adjustbox}
\end{table}
Table~\ref{tab: MMHPSD} presents a comparison of three event representations tested on the MMHPSD dataset: 2D event frames, our 3D rasterized event point clouds (RasEPC), and our 3D decoupled event voxels (DEV) († denotes our reimplementation).
On such a more challenging real-world dataset, under the ResNet18 backbone, the use of DEV representation reduces the MPJPE index from $3.94$ to $2.53$, an improvement of $35.8\%$. 
The DEV-Pose (DHP19) with only $0.91M$ parameters has higher accuracy in key point prediction than the ResNet50 with 2D event frame representation ($3.28$ pixels \textit{vs.} $3.82$ pixels).
Our RasEPC representation consistently delivers excellent real-time estimation accuracy on edge computing platforms, even with more complex real-world datasets. 
Its higher accuracy compared to SPHP (Sparse Convolution)~\citep{lin2023sparse} in 2D event frame representation underscores the importance of preserving temporal dimension information in event data processing.

\begin{table}[!t]
   %
   \caption{\textbf{In-the-wild event-based HPE on the EV-3DPW dataset.}}
   \label{tab:compare_3DPW}
   \centering
   \begin{adjustbox}{width=0.5\textwidth,center}
   \renewcommand\arraystretch{1.2}{\setlength{\tabcolsep}{8mm}{\begin{tabular}{clcccc}
            \toprule
            Input & Method & MPJPE$_{2D}$ \\
            \midrule
            
            \multirow{3}{*}{2D RGB Frames} 
            & DHP19$_{RGB}$~\citep{xiao2018simbase} & 45.17 \\
            & MobileHP-S$_{RGB}$~\citep{choi2021mobilehumanpose} & \textbf{22.22} \\
            & Pose-ResNet18$_{RGB}$~\citep{xiao2018simbase} & 27.50 \\
            
            \midrule
            
            \multirow{4}{*}{2D Event Frames} 
            & DHP19†~\citep{calabrese2019dhp19} & 36.15 \\
            & MobileHP-S†~\citep{choi2021mobilehumanpose} & \textbf{16.95} \\
            & Pose-ResNet18†~\citep{xiao2018simbase} & 17.87 \\
            & Pose-ResNet50†~\citep{xiao2018simbase} & 17.42 \\
            
            \midrule

            \multirow{3}{*}{3D Event Point Cloud}
            & RasEPC (PointNet~\citep{qi2017pointnet}) & 20.65 \\
            & RasEPC (DGCNN~\citep{wang2019dgcnn})) & \textbf{19.98} \\    
            & RasEPC (Point Transformer~\citep{zhao2021pointtrans})) & 20.39 \\    
            \midrule
            
            \multirow{2}{*}{3D Decoupled Event Voxel}
            & DEV-Pose (DHP19) & 28.79 \\      
            & DEV-Pose (ResNet18) & \textbf{15.68} \\ 

            \bottomrule
         \end{tabular}}}
     \end{adjustbox}
\end{table}

%
\noindent\textbf{In-the-Wild Results on the EV-3DPW Dataset.}  
As a derived dataset simulated from the video HPE dataset, the EV-3DPW dataset comprises two distinct data types for in-the-wild scenarios: RGB images and events.
We conduct experiments using both modalities on the same networks, as shown in Table~\ref{tab:compare_3DPW}.
Event frame input outperforms RGB input by responding only to intensity changes, reducing the impact of lighting variations and color similarities between the target and background.
Notably, the 3D RasEPC approach achieves good performance with low latency.
On the other hand, DEV-Pose (ResNet18) reduces MPJPE from $17.87$ to $15.68$ compared to Pose-ResNet18† by a clear margin of $12.3\%$.
These quantitative results demonstrate that DEV representation is superior to the commonly employed 2D representation for accurately estimating human poses in challenging outdoor and intricate environments.
We further present qualitative results in Fig.~\ref{compare}. 
In some difficult scenes, the network working with RGB images fails, while the event frame method delivers better results. 
In the supplementary material, we include results on the noisy-included EV-3DPW dataset. 
Despite the noise, the event frame method still outperforms RGB under the same backbone due to the significant differences between train and test splits in the outdoor dataset, such as background complexity, action categories, and clothing color. 
Events capture only brightness changes, leading to stronger generalization.  
Our RasEPC representation maintains high accuracy with low latency, and the DEV representation achieves higher accuracy compared to 2D event frame representation under the same backbone.

\begin{figure*}[!t]
\centering
\includegraphics[width=1\linewidth]{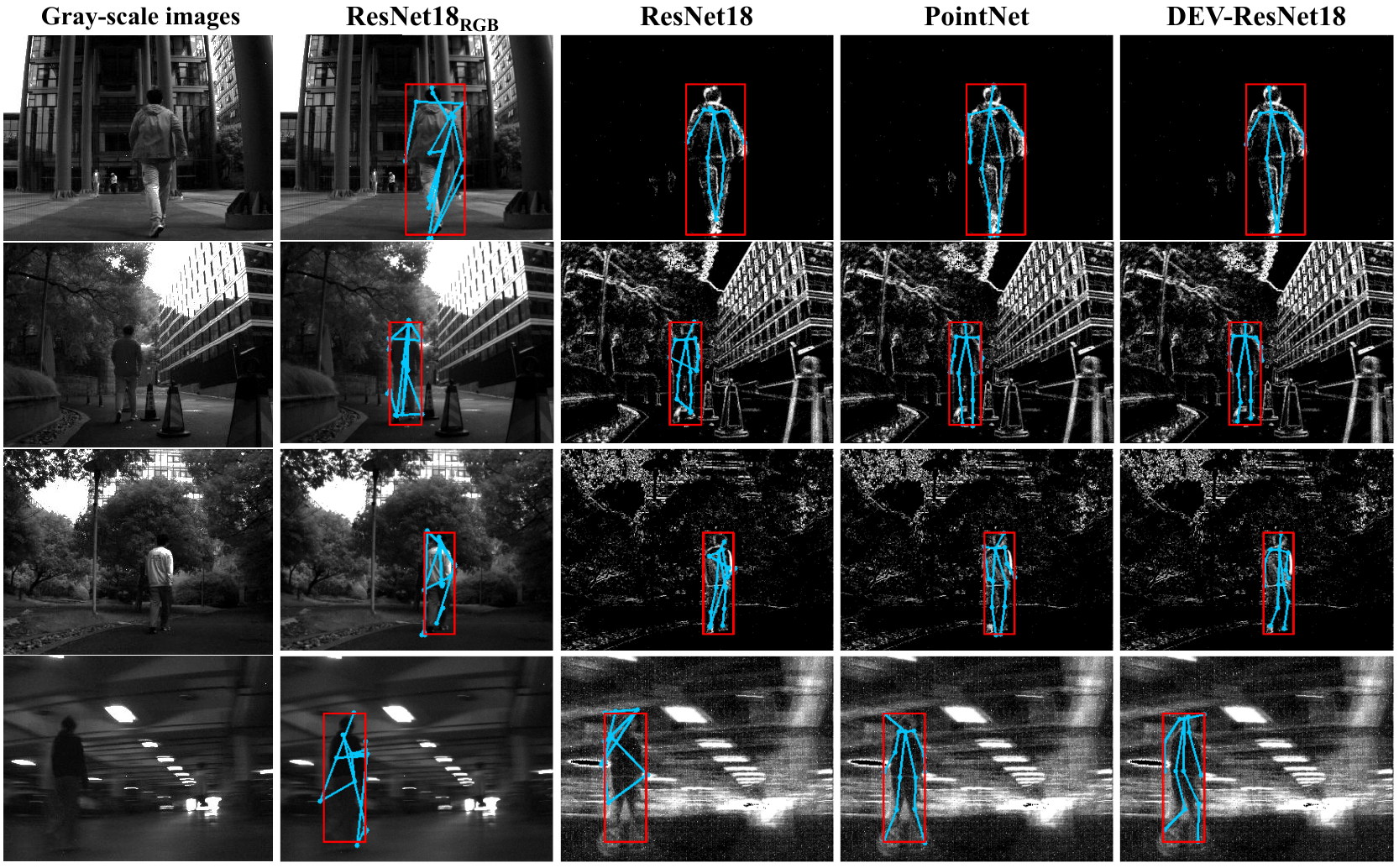}
\vskip-2ex
\caption{Qualitative comparison of different methods in outdoor gray-scale image sequences and event streams captured by our event camera. The human bounding boxes are estimated by the pre-trained YOLOv3 model~\citep{redmon2018yolov3} using MMDetection~\citep{chen2019mmdetection}. Our two 3D event representation methods yield reliable estimates in street scenes and basements, which means stronger generalization ability in the real world.}
\label{compare_Driving}
\vskip-3ex
\end{figure*}

\begin{figure}[!t]
\centering
\includegraphics[width=0.95\linewidth]{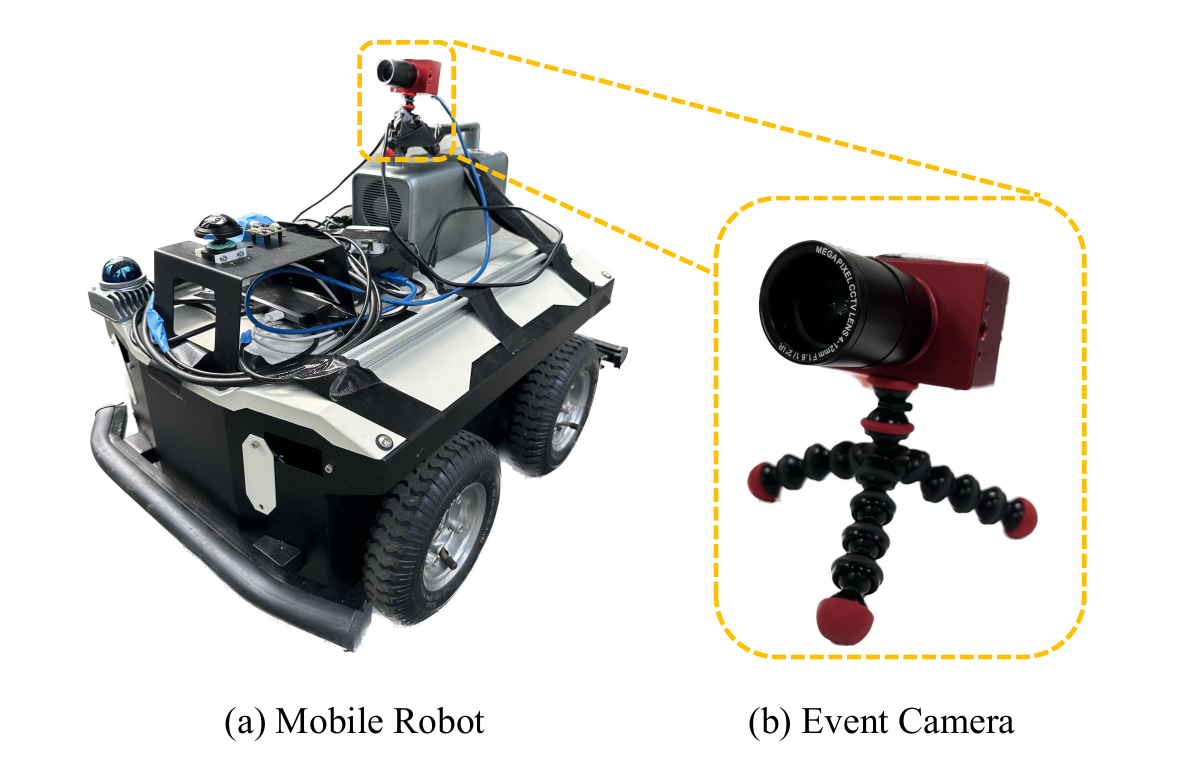}
\vskip-2ex
    \caption{(a) Our outdoor mobile robot is equipped with an event camera (DAVIS-346) and a laptop. (b) Event camera for capturing outdoor aligned grayscale frames and event information.}
\label{MobileRobot}
\vskip-3ex
\end{figure}

Both 3D event representations harness temporal information effectively, highlighting their potential advantages in tasks related to human behavior analysis, especially in predicting keypoints for static limbs.
Such comprehensive descriptions that incorporate temporal dimension information further confirm the practicality of using event cameras for human behavior analysis, particularly in complex outdoor environments with unpredictable motion.

\noindent\textbf{Results Visualization and Real-World Estimations.}
We further qualitatively compare the synthetic driving scene dataset EV-JAAD without ground truth and the collected event streams to verify the generalization ability of the 3D event representation.
As shown in Fig.~\ref{compare_EVJAAD}, RasEPC and DEV representations both give high-quality human pose estimation in new unfamiliar scenes, demonstrating their strong generalization ability.
As for the RGB image method, it suffers from lighting conditions and background changes. 
The event frame method, which discards temporal information entirely, appears less suitable for handling unfamiliar scenes.

To further investigate the practical performance of the proposed 3D event representation solution on real data, we collect aligned grayscale frames and event streams of street scenes and basements, as illustrated in Fig.~\ref{compare_Driving}.
In practice, we install an event camera with a resolution of $346{\times}260$ on top of a mobile robot (see Fig.~\ref{MobileRobot}), which traverses streets and garages under remote control.
Although the robot’s viewing perspective and movement mode are significantly different from that of the handheld camera used in EV-3DPW for training, both 3D event representations provide reliable estimation. 
For other methods, estimating directly on real-world data results in significant challenges.
In dimly lit environments such as garages (last row of Fig.~\ref{compare_Driving}), event-based information is particularly advantageous over traditional image-based methods. 
However, the event frame approach fails to accurately estimate any joint points, while both three-dimensional event representations yield considerably more reliable results. 
Correct estimation of pedestrian poses in garage scenarios holds significant importance, especially in applications like smart parking.

In summary, our solution demonstrates superior performance compared to RGB image-based and event-frame methods using the same backbones for human pose estimation, showing excellent synthetic-to-real generalizability.

\section{Limitations}
While we have demonstrated the effectiveness of 3D event representations for human pose estimation, there are certain limitations of our study.
Firstly, we have primarily focused on their application in HPE without exploring their potential in other event-related tasks, which is an interesting avenue for future research.
Secondly, in multi-person scenarios, our method are built in a top-down fashion, relying on preprocessed human bounding boxes without end-to-end optimization, which leaves room for further exploration and refinement of our approach.

\section{Conclusion}
In this work, we look into event-based human pose estimation from a novel perspective of 3D event representations.
Existing event-frame-based methods accumulate event signals into synchronized frames, undermining the natural high temporal resolution of events. 
To preserve these high-resolution features, we explore representing events in the three-dimensional space of $[x, y, t]$.
The proposed idea is implemented with two novel representations, namely the rasterized event point cloud representation and the decoupled event voxel representation.
We further introduce EV-3DPW, a public synthetic event point cloud dataset, which facilitates the training and evaluation of event-based HPE models. 
Experiments on the public DHP19 dataset, MMHPSD dataset, and our established EV-3DPW dataset demonstrate that event point cloud representation with three known point-wise backbones attains good trade-offs between speed and accuracy. 
Evidently, the decoupled event voxel representation is compatible with well-known 2D CNN backbones, which significantly improves the accuracy of human pose estimation while ensuring computational efficiency. 
In general, sparse point cloud input (RasEPC) is more suitable for real-time applications on edge computing platforms, whereas the DEV representation, which includes neighborhood information, achieves more accurate key point estimation than the 2D event frame representation using the same backbone.
Both 3D event representations demonstrate strong generalizability in unseen scenarios.

In the future, we look forward to further exploring the adaptability of the 3D event representation for other downstream tasks related to human behavior.
Precisely, we aim to explore other event-based human behavior understanding tasks, such as action recognition and gait recognition. 
Furthermore, we plan to leverage synthetic in-the-wild datasets to achieve end-to-end HPE for multiple people. 
We also intend to collect a real outdoor multi-person dataset to provide a benchmark for the quantitative evaluation of event-based human pose estimation in difficult scenes and stimulate new research in this field.

\bibliographystyle{model2-names}
\bibliography{refs_v2}

\noindent\textbf{Visualization and Analysis of Features from Different Event Representations.} 
To further explore the differences among various event representation methods, we compare and analyze the Event Frame, HATS~\citep{sironi2018hats}, Ev-FlowNet~\citep{zhu2018ev}, EST~\citep{gehrig2019end}, and our proposed DEV representation.
In Ev-FlowNet~\citep{zhu2018ev}, events are encoded as time surfaces across four channels.
The first two channels capture the number of positive and negative events at each pixel, while the last two channels record the timestamps of the most recent positive and negative events at each pixel. 
Although this method retains the latest timestamps, it discards earlier ones, leading to a loss of temporal information.
Similarly, HATS~\citep{sironi2018hats} represents events using a histogram of average time surfaces but still discards temporal information.
On the other hand, representing events as a voxel grid~\citep{zhu2019unsupervised} involves discretizing the time dimension into $B$ bins and treating the time channel as the $C$ channel in the network. 
Unfortunately, this approach restricts interaction between different bins, limiting the network’s ability to fully utilize temporal information.
In contrast, our proposed representations are designed to overcome these limitations by preserving and leveraging more comprehensive temporal information, resulting in better performance.

\begin{figure*}[!t]
\centering
\includegraphics[width=1.0\linewidth]{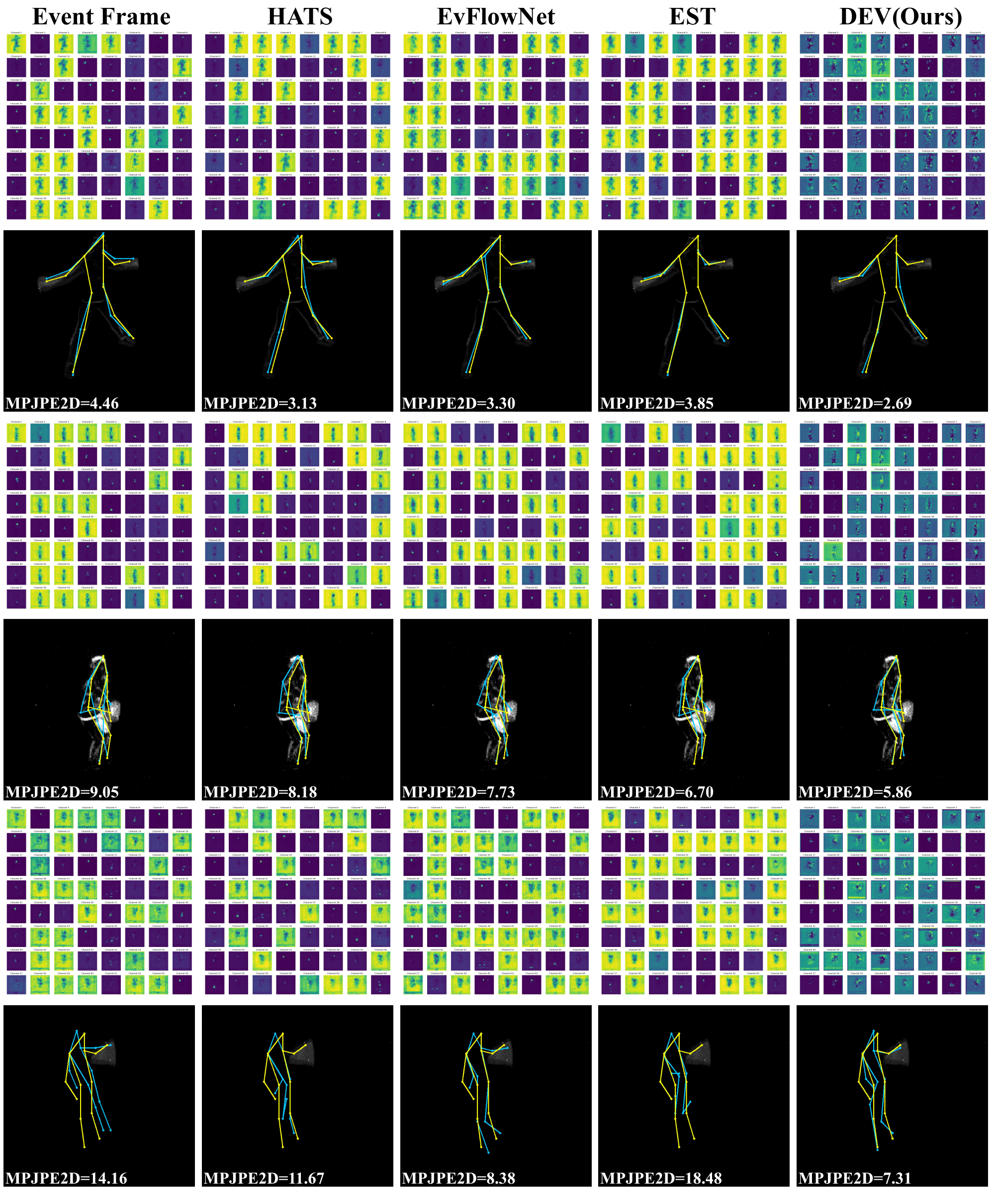}
\caption{Visualization of feature maps and key point estimation results.}
\label{Feature Visualization}
\end{figure*}

As illustrated in Figure \ref{Feature Visualization}, we present the feature maps visualized on a per-channel basis, along with their corresponding estimation results, using the DHP19 backbone on the DHP19 dataset~\citep{calabrese2019dhp19}.
The feature maps produced by the Event Frame representation appear blurred and scattered, with many channels displaying similar features and lacking distinct structures. 
This suggests that the method struggles to effectively capture spatial and temporal information, primarily due to the limited richness of information inherent in event frames.
The HATS~\citep{sironi2018hats} representation offers more structured feature maps but still lacks finer details.
Although this method captures some global information by generating histograms of average time surfaces, it sacrifices a significant portion of temporal information, resulting in feature maps that are somewhat inadequate for detailed analysis.
EvFlowNet~\citep{zhu2018ev} produces more structured and clear feature maps, especially in representing human contours.
While it performs well in keypoint estimation, it discards earlier timestamps, which may limit its effectiveness in handling long sequences or complex motions, indicating potential areas for improvement.
The EST~\citep{gehrig2019end} method employs a multi-layer perceptron (MLP) to map event points into feature representations in a learnable, end-to-end optimized manner. 
The resulting feature maps display finer and richer structural information, leading to good performance in key point estimation.
The complexity introduced by MLP-based processing can make the network harder to train, especially on datasets with limited diversity or size, potentially leading to suboptimal performance compared to simpler representations like event histograms~\citep{sironi2018hats} or time surfaces~\citep{zhu2018ev}.
In contrast, our proposed DEV representation demonstrates clear and rich structural features in its maps. 
The sharp contour edges indicate that the network is effectively capturing and representing object boundaries and areas near key points. 
Additionally, by incorporating fine-grained temporal information from both the $xt$ and $yt$ projections, the DEV representation is able to focus on more relevant feature channels (as indicated by brighter activations), thus avoiding temporal feature mixing that can blur distinctions. 
This channel-wise selectivity, absent in other representations, enhances the DEV approach's ability to accurately differentiate between key points, improving the precision of keypoint localization.

\begin{figure*}[!t]
\centering
\includegraphics[width=1\linewidth]{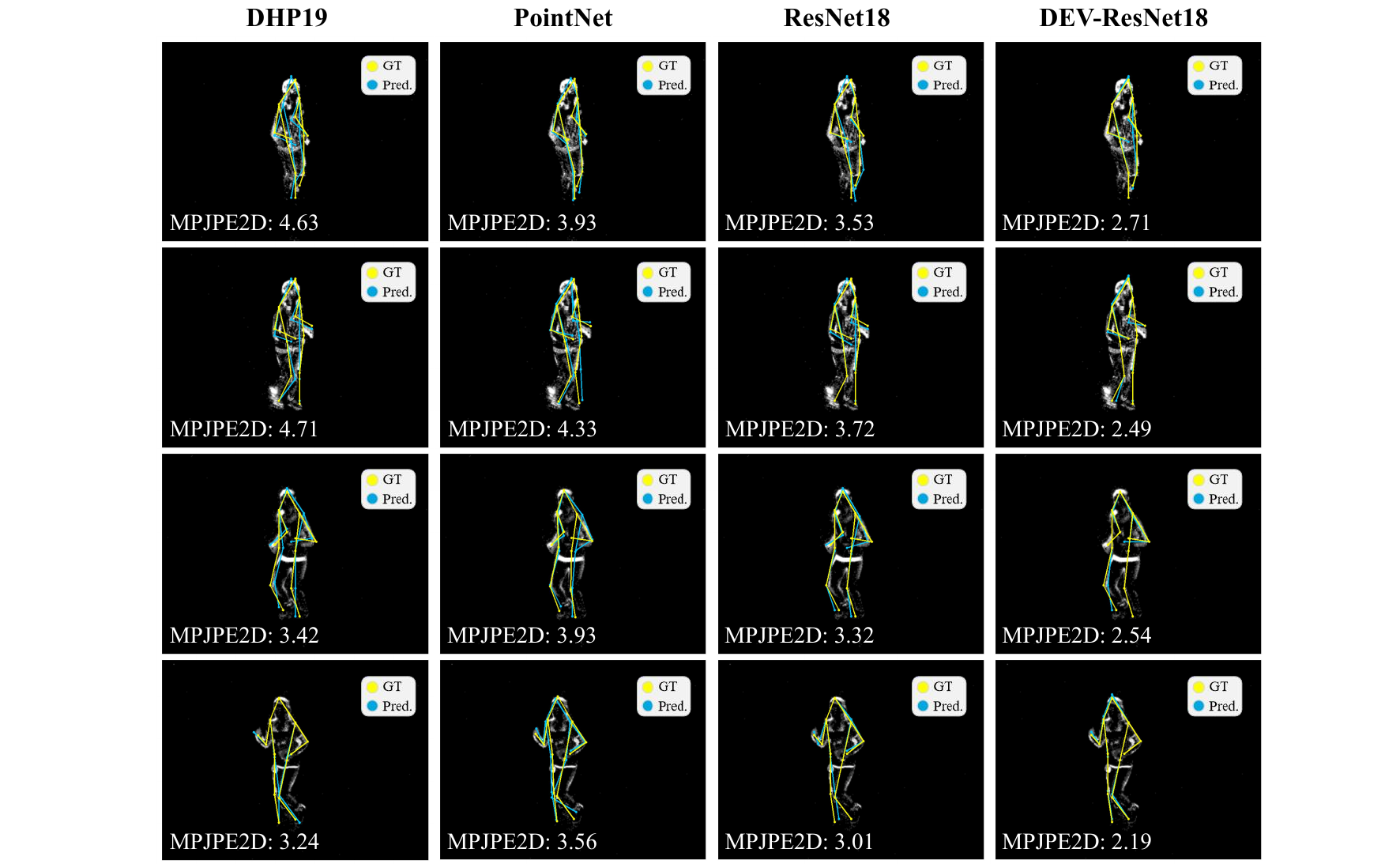}
\vskip-3ex
\caption{2D results visualization on the DHP19 test dataset for different models (yellow for ground truth, blue for prediction).} 
\label{Supp_DHP19_1}
\end{figure*}

\begin{figure*}[!t]
\centering
\includegraphics[width=1\linewidth]{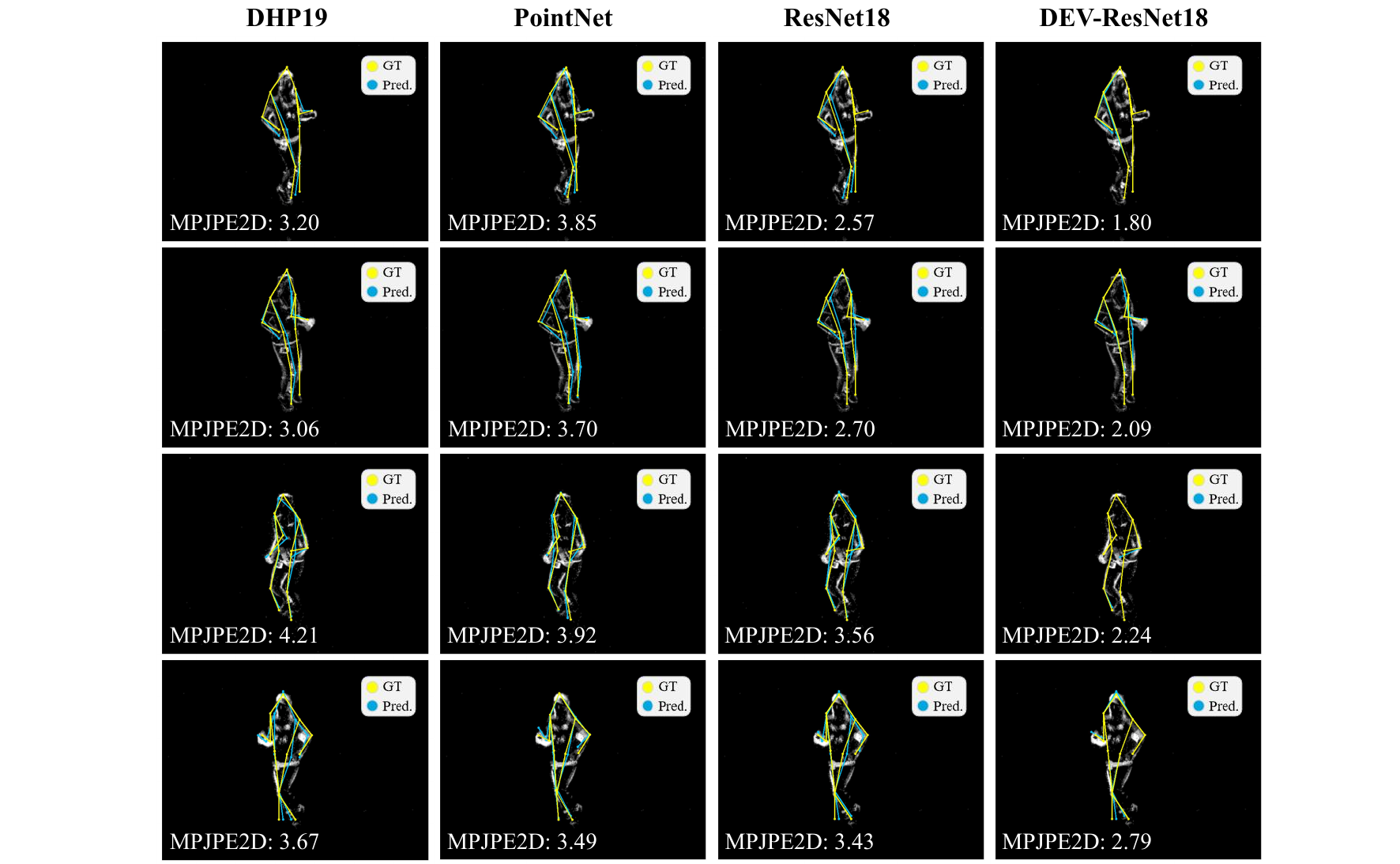}
\vskip-3ex
\caption{Additional results by DHP19~\citep{calabrese2019dhp19}, our PointNet~\citep{qi2017pointnet}, Pose-ResNet18†~\citep{xiao2018simbase}, and our DEV-Pose (ResNet18) on the DHP19 test dataset.} 
\label{Supp_DHP19_2}
\end{figure*}

\noindent\textbf{More Visualizations on the indoor DHP19 dataset.} We provide qualitative results on the DHP19~\citep{calabrese2019dhp19} dataset through different event representation methods, as seen in Fig.~\ref{Supp_DHP19_1} and Fig.~\ref{Supp_DHP19_2}. Our study reveals that the proposed event-point-cloud-based approach demonstrates increased robustness in managing static limbs compared to the original DHP19 model. This is particularly evident when limbs remain static during motion, such as the right calf in the first row of Fig.~\ref{Supp_DHP19_1} and the legs in the final row of Fig.~\ref{Supp_DHP19_2}. In these instances, a scarcity of generated events can lead to the invisibility of certain parts in the accumulated event frames. However, our event point cloud method retains these sparse events, allowing for more effective processing by our point-wise backbone. This preservation is likely a key factor in our method's improved accuracy in predicting keypoints for static limbs. Furthermore, when utilizing the same ResNet18 backbone, the accuracy of our DEV representation shows significant enhancement over the conventional event frame-based approach.

\begin{figure*}[!t]
\centering
\includegraphics[width=1.0\linewidth]{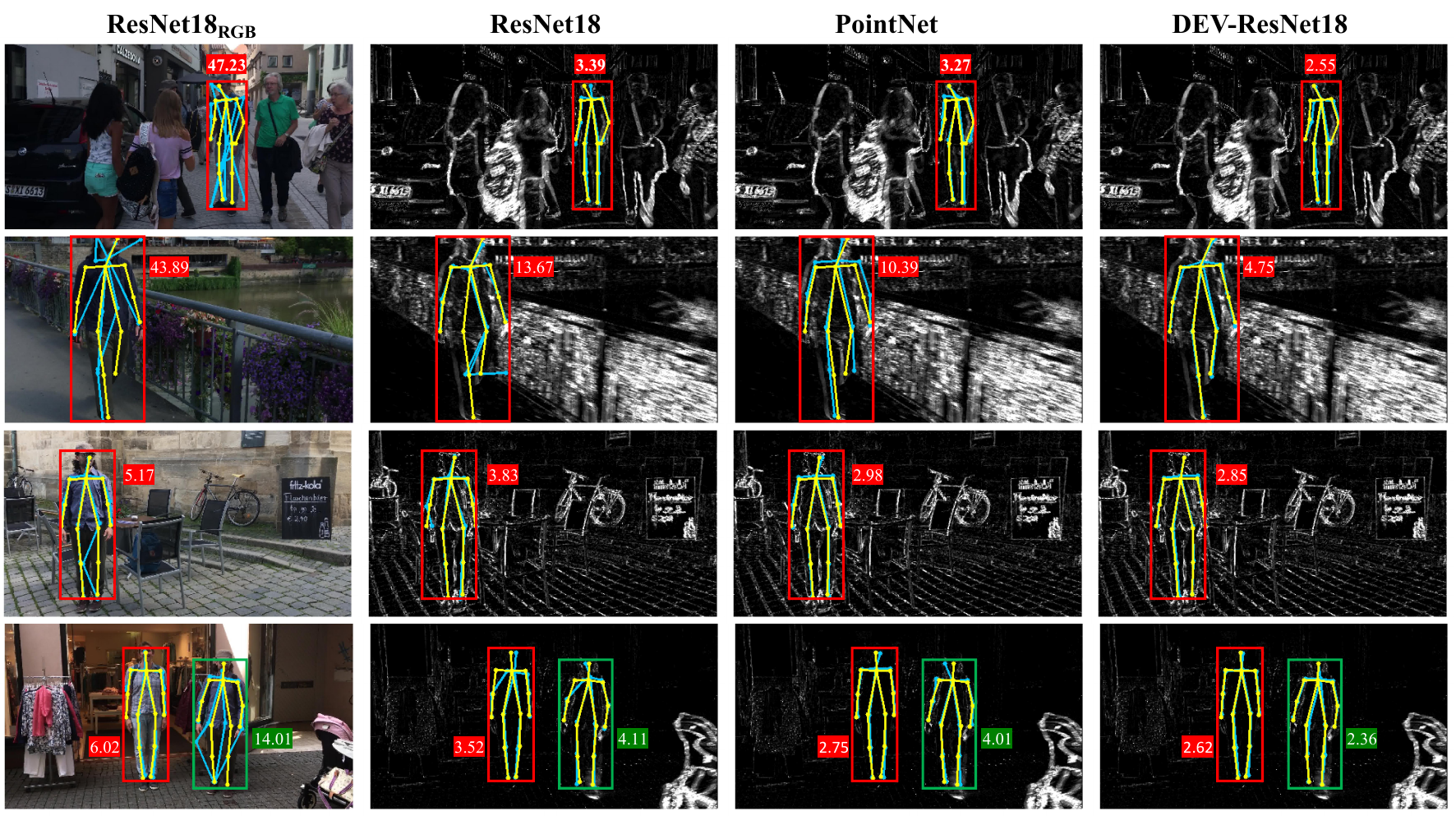}
\caption{Additional results by DHP19~\citep{calabrese2019dhp19}, our PointNet~\citep{qi2017pointnet}, Pose-ResNet18†~\citep{xiao2018simbase}, and our DEV-Pose (ResNet18) on EV-3DPW test dataset.}
\label{Supp_3DPW}
\end{figure*}

\noindent\textbf{More Visualizations on the outdoor EV-3DPW dataset.} Further results on the EV-3DPW dataset in outdoor settings are showcased in Fig.~\ref{Supp_3DPW}. In alignment with observations noted in the main text, we find that methods dependent on RGB images fall short due to considerable divergences between the training and testing environments. These discrepancies encompass factors such as background complexity, color variations, and occlusion. In contrast, the three methodologies utilizing event data deliver more dependable estimations. This enhanced performance is attributable to their proficiency in capturing differential and motion information, proving particularly effective in varied and complex outdoor scenarios. Among these event-based approaches, our DEV-Pose method demonstrates the most superior performance. This is primarily attributed to its effective modeling of local relationships within spatio-temporal neighborhoods, further enhancing its reliability and accuracy in complex outdoor scenarios.

\noindent\textbf{Quantitative results on EV-3DPW (Noise) dataset.}
To further explore the effect of the EV-3DPW dataset under more realistic conditions, we conduct additional experiments on the EV-3DPW (Noise) dataset generated by DVS-Voltmeter~\citep{lin2022dvsvoltmeter}. 
The results are shown in Table \ref{tab:compare_3DPW_Noise}.
These findings are similar to those obtained on the noise-free dataset.  
This consistency is likely due to the significant gap between the train and test splits of the 3DPW dataset, which includes variations in background complexity, action type, clothing color, etc., providing a substantial advantage to event data that only responds to intensity changes. 
Therefore, the key point accuracy estimated by the three event-based schemes is generally better than that of RGB input. 
While the RasEPC representation can be estimated in real-time on edge computing platforms with comparable accuracy, the DEV representation still achieves the highest accuracy.

\begin{table}[!t]
   %
   \caption{\textbf{In-the-wild event-based HPE on the EV-3DPW (Noise) dataset.}}
   \label{tab:compare_3DPW_Noise}
   \centering
   \begin{adjustbox}{width=0.5\textwidth,center}
   \renewcommand\arraystretch{1.2}{\setlength{\tabcolsep}{8mm}{\begin{tabular}{clcccc}
            \toprule
            Input & Method & MPJPE$_{2D}$ \\
            \midrule
            
            \multirow{3}{*}{2D RGB Frames} 
            & DHP19$_{RGB}$~\citep{xiao2018simbase} & 46.08 \\
            & MobileHP-S$_{RGB}$~\citep{choi2021mobilehumanpose} & \textbf{19.99} \\
            & Pose-ResNet18$_{RGB}$~\citep{xiao2018simbase} & 23.17 \\
            
            \midrule
            
            \multirow{4}{*}{2D Event Frames} 
            & DHP19†~\citep{calabrese2019dhp19} & 35.2 \\
            & Pose-ResNet18†~\citep{xiao2018simbase} & 17.21 \\
            & Pose-ResNet50†~\citep{xiao2018simbase} & \textbf{17.00} \\
            
            \midrule
            
            \multirow{2}{*}{3D Event Point Cloud}
            & RasEPC (PointNet~\citep{qi2017pointnet}) & 20.92 \\
            & RasEPC (DGCNN~\citep{wang2019dgcnn}) & \textbf{19.60} \\    
            \midrule
            
            \multirow{2}{*}{3D Decoupled Event Voxel}
            & DEV-Pose (DHP19) & 28.63 \\      
            & DEV-Pose (ResNet18) & \textbf{16.26} \\ 

            \bottomrule
         \end{tabular}}}
     \end{adjustbox}
\end{table}

\end{document}